\documentclass[letterpaper]{article} 
\usepackage{aaai24}  
\usepackage{times}  
\usepackage{helvet}  
\usepackage{courier}  
\usepackage[hyphens]{url}  
\usepackage{graphicx} 
\usepackage{natbib}  
\usepackage{caption} 
\usepackage{algorithm}
\usepackage{algorithmic}
\usepackage{amsmath}
\usepackage{newfloat}
\usepackage{listings}
\DeclareCaptionStyle{ruled}{labelfont=normalfont,labelsep=colon,strut=off} 
\floatstyle{ruled}
\newfloat{listing}{tb}{lst}{}
\floatname{listing}{Listing}
\usepackage{amsfonts}
\usepackage{amssymb}
\usepackage{graphicx}
\usepackage{verbatim}
\usepackage{xcolor}
\usepackage{tabularx}
\usepackage{booktabs}
\usepackage{lscape}
\usepackage{caption}
\usepackage{subcaption}
\usepackage{xcolor}
\usepackage{booktabs}
\usepackage{multirow}
\usepackage{siunitx}
\usepackage{varioref}
\usepackage{mathabx}
\usepackage{textcomp}
\usepackage{cleveref}
\usepackage{paralist}
\usepackage{mathtools}
\usepackage{tikz}
\usepackage{multirow}
\usepackage{colortbl}
\usepackage{array}
\usepackage{booktabs}

\renewcommand{\vec}[1]{\boldsymbol{#1}}

\title{Interactive Hyperparameter Optimization in Multi-Objective Problems\\via Preference Learning}
\author{
    Joseph Giovanelli\textsuperscript{\rm 1}, Alexander Tornede\textsuperscript{\rm 2}, Tanja Tornede\textsuperscript{\rm 2}, Marius Lindauer\textsuperscript{\rm 2}
}
\affiliations{
    \textsuperscript{\rm 1}Alma Mater Studiorum --- University of Bologna\\
    \textsuperscript{\rm 2}Institute of Artificial Intelligence, L3S Research Center, Leibniz University Hannover\\
    j.giovanelli@unibo.it, \{a.tornede, t.tornede, m.lindauer\}@ai.uni-hannover.de
}

\begin{document}

\maketitle

\begin{abstract}
Hyperparameter optimization (HPO) is important to leverage the full potential of machine learning (ML). 
In practice, users are often interested in multi-objective (MO) problems, i.e., optimizing potentially conflicting objectives, like accuracy and energy consumption.
To tackle this, the vast majority of MO-ML algorithms return a Pareto front of non-dominated machine learning models to the user.
Optimizing the hyperparameters of such algorithms is non-trivial as evaluating a hyperparameter configuration entails evaluating the quality of the resulting Pareto front. 
In literature, there are known indicators that assess the quality of a Pareto front (e.g., hypervolume, R2) by quantifying different properties (e.g., volume, proximity to a reference point). However, choosing the indicator that leads to the desired Pareto front might be a hard task for a user. In this paper, we propose a human-centered interactive HPO approach tailored towards multi-objective ML leveraging preference learning to extract desiderata from users that guide the optimization.
Instead of relying on the user guessing the most suitable indicator for their needs, our approach automatically learns an appropriate indicator.
Concretely, we leverage pairwise comparisons of distinct Pareto fronts to learn such an appropriate quality indicator.
Then, we optimize the hyperparameters of the underlying MO-ML algorithm towards this learned indicator using a state-of-the-art HPO approach.
In an experimental study targeting the environmental impact of ML, we demonstrate that our approach leads to substantially better Pareto fronts compared to optimizing based on a wrong indicator pre-selected by the user, and performs comparable in the case of an advanced user knowing which indicator to pick.

\end{abstract}

\section*{Introduction}
\label{sec:introduction}
Practical applications of machine learning often call for the optimization of more than one loss or objective function, called multi-objective machine learning (MO-ML) \citep{jin-momlbook06a} borrowing from the notion of multi-objective optimization (MO) \citep{deb-ds16,gunantara-ce18}. For example, instead of only focusing on the performance of a model, its energy consumption is becoming more and more important in various domains such as edge computing, but also in general, sparked by efforts in the area of green artificial intelligence (Green AI) \citep{schwartz-arxiv19a,wynsberghe-aiethics21a}. Comparing two ML models with respect to several objectives is non-trivial and most approaches solve this problem by returning a set of solutions that cannot be improved further without going to the expense of at least one of the objectives, called the Pareto front.

Naturally, as standard ML algorithms, MO-ML algorithms expose hyperparameters controlling their learning behavior and thus, also the resulting Pareto front. To unleash the full potential of the MO-ML algorithms, these hyperparameters should be optimized with adequate methods. Unfortunately, optimizing the hyperparameters of such MO-ML algorithms is challenging for a user with standard hyperparameter optimization (HPO) \citep{feurer-automlbook19a,bischl-dmkd23a} approaches, such as SMAC~\citep{hutter-lion11a,lindauer-jmlr22a}, Optuna~\citep{akiba-kdd19a}, Hyperopt~\citep{komer-scipy14a}, HpBandSter~\citep{falkner-icml18a} or SyneTune~\citep{salinas-automl22}, which iteratively evaluate configurations. This is the case, as evaluating a hyperparameter configuration of an MO-ML algorithm involves evaluating the quality of the Pareto front of models returned by the corresponding algorithm. Several so-called (quality) indicators, such as hypervolume \citep{zitzler1999multiobjective}, are used to assess different properties of the shape of the Pareto front and in principle can also be used for the purpose of rating a configuration in the context of HPO. Nevertheless, although a user might have a clear idea about what kind of Pareto front shape they would like to choose their final model from, choosing the quality indicator leading to this Pareto front shape is a challenging task in practice. At the same time, however, correctly configuring the HPO tool with the right loss function, i.e. quality indicator in our case, is crucial to achieve the desired result. 

With this paper, we propose an interactive human-centered HPO \citep{pfisterer-arxiv2019a,souza-ecmlpkdd21a,moosbauer-neurips21a,hvarfner-iclr22a,moosbauer-arxiv22a,francia-fgcs23a,segel-automl23a,mallik-arxiv23} approach for MO-ML algorithms that frees users from choosing a predefined quality indicator suitable for their needs by learning one tailored towards them based on feedback. 
To achieve this, it first learns the desired Pareto front shape from the user in a short interactive session and then starts a corresponding HPO process optimizing towards the previously learned Pareto front shapes. 
Instead of requiring the user to present us with a concrete Pareto front they would favor, we interactively and iteratively present them a few pairs of Pareto fronts asking them for their preferences.
Based on these pairwise comparisons, we leverage methods from the field of preference learning (PL) \citep{furnkranz-plbook10a} to learn a latent utility function serving as a Pareto front quality indicator customized to the user. 
In the subsequent stage, we run a state-of-the-art HPO tool 
instantiated with the learned Pareto front quality indicator to evaluate configurations and the corresponding Pareto fronts. 

In summary, we make the following contributions:
\begin{enumerate}
    \item We propose an interactive approach to learn a Pareto front quality indicator from pairwise comparisons based on a latent utility function with methods from the field of preference learning. This quality indicator is customized to the user and can be learned from a small number of pairwise comparisons. 
    \item We combine the aforementioned learned quality indicator with an HPO tool 
    to provide a full-fledged HPO approach for multi-objective MO-ML algorithms, which frees the user to choose an appropriate Pareto front quality indicator offering less opportunity for a mistake and thus, bad optimization results.
    \item In an experimental case study, we demonstrate that our approach leads to substantially better Pareto fronts compared to optimizing based on a wrong indicator pre-selected by the user, and performs comparable in case of an advanced user knowing which indicator to pick. Thus, our approach makes HPO for MO-ML algorithms substantially more easily and robustly applicable in practice.
\end{enumerate}

\section*{Background}
\label{sec:background}

Since our work is spanned along the dimensions of hyperparameter optimization (HPO), multi-objective optimization (MO), and preference learning (PL), we introduce in the following all of the aforementioned concepts.

\subsection*{Hyperparameter Optimization} \label{ssec:hpo}
HPO formalizes the task of finding a hyperparameter configuration for a machine learning algorithm leading to a well-performing model on a given dataset 
\begin{equation}
    \mathcal{D} = \{(\vec{x}_n, y_n)\}_{n=1}^N \in \mathbb{D} \subset \mathcal{X} \times \mathcal{Y}
\end{equation}
with an instance space $\mathcal{X}$ and a target space $\mathcal{Y}$. 
In addition to the dataset, we are provided with a hyperparameter configuration space $\boldsymbol{\Lambda} = \Lambda_1 \times \dots \times \Lambda_K$ with $K$ hyperparameters, where $\Lambda_k$ is the domain of the $k^{\mathit{th}}$ hyperparameter, and an algorithm $A: \mathbb{D} \times \vec{\Lambda} \rightarrow \mathcal{H}$ which trains a model from the model space $\mathcal{H}$ given a dataset and a hyperparameter configuration. 
Furthermore, we are provided with a loss function $\mathcal{L}: \mathcal{H} \times \mathbb{D} \rightarrow \mathbb{R}$ quantifying how well a given model performs on a given dataset. 
The loss function can be used to assess the quality of a hyperparameter configuration by splitting the original dataset $\mathcal{D}$ into two disjoint datasets $\mathcal{D}_\mathit{train}$ and $\mathcal{D}_\mathit{test}$, where the model is trained only based on $\mathcal{D}_\mathit{train}$ but evaluated with $\mathcal{L}$ on $\mathcal{D}_\mathit{test}$. 
Overall, we seek to find the optimal hyperparameter configuration $\vec{\lambda}^* \in \vec{\Lambda}$ defined as 
\begin{equation}\label{eq:hpo}
    \vec{\lambda}^* \in \arg\min_{\vec{\lambda} \in \vec{\Lambda}} \mathcal{L}\left( A \left( \mathcal{D}_{\mathit{train}}, \vec{\lambda} \right), \mathcal{D}_\mathit{test} \right) \, .
\end{equation}

There exist several approaches to automatically solving the optimization problem defined in \eqref{eq:hpo}, many of which are powered by Bayesian optimization (BO) \citep{frazier-arxiv18a} or evolutionary approaches \citep{bischl-dmkd23a}. 
Most of these techniques internally iteratively evaluate a large set of configurations based on their true estimated loss. 
To this end, they split up a so-called validation dataset $\mathcal{D}_\mathit{val}$ from the training dataset $\mathcal{D}_\mathit{train}$ to estimate the loss $\mathcal{L}(A(\mathcal{D}_{\mathit{train}}, \vec{\lambda}), \mathcal{D}_\mathit{test})$ of a configuration $\vec{\lambda}$ by $\mathcal{L}(A(\mathcal{D}_{\mathit{train}}, \vec{\lambda}), \mathcal{D}_\mathit{val})$ and avoid a biased overfit to $\mathcal{D}_\mathit{test}$.

\subsection*{Multi-Objective Machine Learning}
\label{ssec:moo}

When it comes to learning a machine learning model with more than one (possibly conflicting) objective or loss function $\mathcal{L}_1, \dots, \mathcal{L}_M$ in mind, comparing the quality of models becomes difficult. For instance, in the context of Green AI, searching for a model with higher performance usually leads to one with higher power consumption, and vice versa. Thus, in multi-objective ML, learning algorithms leverage the concept of dominance formalizing that one model $h_1 \in \mathcal{H}$ is better than another $h_2 \in \mathcal{H}$, if $h_1$ performs better in at least one of the objectives while not performing worse than $h_2$ in the remaining ones. Yet, this leaves us with a set of non-dominated models, which are indistinguishable with respect to this dominance idea. These models form a so-called Pareto front $P_{\mathcal{D}_{\mathit{val}}}(H)$ evaluated on dataset $\mathcal{D}_{\mathit{val}}$ for a given set of models $H \subset \mathcal{H}$. Formally, a Pareto front is defined as 

\begin{equation}
\label{eq:pareto_def}
    P_{\mathcal{D}_{\mathit{val}}}(H) = \left\{ h \,\, \begin{array}{|l} 
    h \in H, \nexists h' \in H : \\
    \forall m \in \{1, \dots, M\} : \\
        \mathcal{L}_m(h',\mathcal{D}_{\mathit{val}}) \leq \mathcal{L}_m(h,\mathcal{D}_{\mathit{val}}), \\
    \exists j \in \{1, \dots, M\} : \\
        \mathcal{L}_j(h',\mathcal{D}_{\mathit{val}}) < \mathcal{L}_j(h, \mathcal{D}_{\mathit{val}})
    \end{array}\right\} \, .
\end{equation} 

Due to the problem of further distinguishing non-dominated models among each other, MO-ML algorithms usually resolve to returning the complete Pareto front of models instead of a single model. Thus, formally, the signature of an algorithm changes to $A: \mathbb{D} \times \vec{\Lambda} \rightarrow 2^{\mathcal{H}}$. As a consequence, the evaluation of each hyperparameter configuration of such an MO-ML algorithm also involves quantifying the quality of a Pareto front of models instead of a single model returned by the algorithm, yielding a much more difficult problem. As such, the signature of our HPO loss function used in Eq.~\ref{eq:hpo} also needs to change to $\mathcal{L}: 2^{\mathcal{H}} \times \mathbb{D} \rightarrow \mathbb{R}$ and thus can no longer be instantiated with simple loss functions such as accuracy. 

One way of assessing the quality of a Pareto front and thus instantiating the HPO loss function $\mathcal{L}$ in this case, are so-called Pareto front quality indicators~\citep{audet-ejor21}. 
These can be categorized into so-called external and internal indicators where external indicators measure the convergence of a Pareto front to the optimal one, i.e., the one that the user prefers. 
Yet, in real-case problems, computing the entire Pareto front space is unfeasible and there is no way to describe the desiderata beforehand.
In contrast, internal indicators assess the Pareto front quality by measuring specific characteristics. 
A very common indicator is called hypervolume \citep{zitzler1999multiobjective} quantifying the volume occupied by the Pareto front w.r.t. a reference point.
Other indicators evaluate factors such as uniformity of solution distribution (e.g., spacing indicator \citep{schott1995fault}), range of values covered by the Pareto front (e.g., maximum spread indicator \citep{zitzler2000comparison}), or proximity to specific threshold points (e.g., R2 indicator \citep{hansen1994evaluating}).
Even though internal indicators can be effectively used as a loss function in HPO of MO-ML algorithms to quantify the quality of Pareto front and thus of a configuration, choosing the measure leading to a Pareto front which has a desired shape requires deep expert knowledge and thus, is a hard task for the user. In particular, a user has to map their implicit desiderata for the Pareto front shape to properties of the quality indicators without a concrete way of tailoring the HPO approach to these desiderata.

\subsection*{Preference Learning}
\label{ssec:preference_learning_related}

Preference learning (PL) \citep{furnkranz-plbook10a} is a subfield of machine learning dealing with the problem of learning from different forms of preferences. Although, PL in general comprises a large set of learning problems, we  focus on the object ranking problem \citep{furnkranz-plbook10a} and in particular of learning to rank objects based on a given set of pairwise preferences. 

In the object ranking problem, we consider a space of objects $\mathcal{O}$, where each object $o \in \mathcal{O}$ is represented as a feature vector $\vec{f}_o$. Such an object can be an item, a document, or (in our case) a Pareto front. Preferences among two objects $o_i, o_j \in \mathcal{O}$ are denoted as $o_i \succ o_j$ indicating that object $o_i$ is preferred over $o_j$. The corresponding space of pairwise rankings over $\mathcal{O}$ is denoted as $\mathcal{R}(\mathcal{O})$. Then, given a dataset of the form $\mathcal{U} = \{o_{i,1} \succ o_{i,2}\}_{i=1}^U$, the goal is to learn a function $f: \mathcal{O} \times \mathcal{O} \rightarrow \mathcal{R}(\mathcal{O})$ that, given two objects, returns the correct pairwise ranking of these two objects. 

Most approaches to object ranking work by learning a utility function $u: \mathcal{O} \rightarrow \mathbb{R}$ returning a utility score for an object that can be used to create a ranking of two objects by sorting them according to their utility score. 

\section*{Related Work}\label{sec:background_work}
In the following, we give a short overview of related work in the areas of human-centered HPO/AutoML and the usage of preference learning in AutoML-related fields.

The field of human-centered HPO/AutoML has gained increasing traction in the last years with approaches targeted at explaining the hyperparameter optimization process to increase the trust in automated tools \citep{pfisterer-arxiv2019a,moosbauer-neurips21a,moosbauer-arxiv22a,segel-automl23a}. This also includes concrete tools developed to help a user interpret the results such as XAutoML \citep{zoller-arxiv22a} or DeepCave \citep{sass-realml22a}. 

Motivated from a similar stance that the user should be put back into the loop of the AutoML process to a certain extent, \citet{francia-fgcs23a} propose to leverage structured argumentation to semi-automatically constrain search spaces based on user input. Further going into this direction and most similar to our approach is the work by \citet{kulbach-ecai20a} building upon the observation that users are often unable to concretely configure a loss function in an AutoML tool fitting for their problem at hand. They suggest to learn a loss function customized to the user as a scalarization over several frequently used standard loss functions via PL methods. Our work differs from theirs in two main aspects: 
\begin{inparaenum}[(1)]
    \item First, we consider tuning the hyperparameters of MO-ML algorithms, while they try to find complete AutoML pipelines with standard classification/regression algorithms. Hence, there is not only a difference in the considered meta-problem (AutoML vs. HPO), but also in the types of machine learning algorithms to be composed / tuned (pipelines of standard ML algorithms vs MO-ML algorithms).
    \item Second, we demonstrate pairwise comparisons of Pareto fronts to the user compared to pairwise comparisons of feature vectors, targets and predictions presented by \citet{kulbach-ecai20a}. We believe that our comparisons are much easier to make for a user.
    \item Moreover, the Pareto front quality indicator learned by us is not constrained to be a scalarization of existing loss functions or quality indicators. Instead, we can, in principle, leverage any object ranking approach that allows for the extraction of a utility function.
\end{inparaenum}

Lastly, our approach is not to be confused with multi-objective HPO \citep{moraleshernandez-arxiv21a,karl-arxiv22a}, where HPO problem itself features multiple loss functions to be optimized, but the ML algorithm itself only returns a single solution.

\begin{figure*}[!ht]
\centering
\includegraphics[width=2.2\columnwidth]{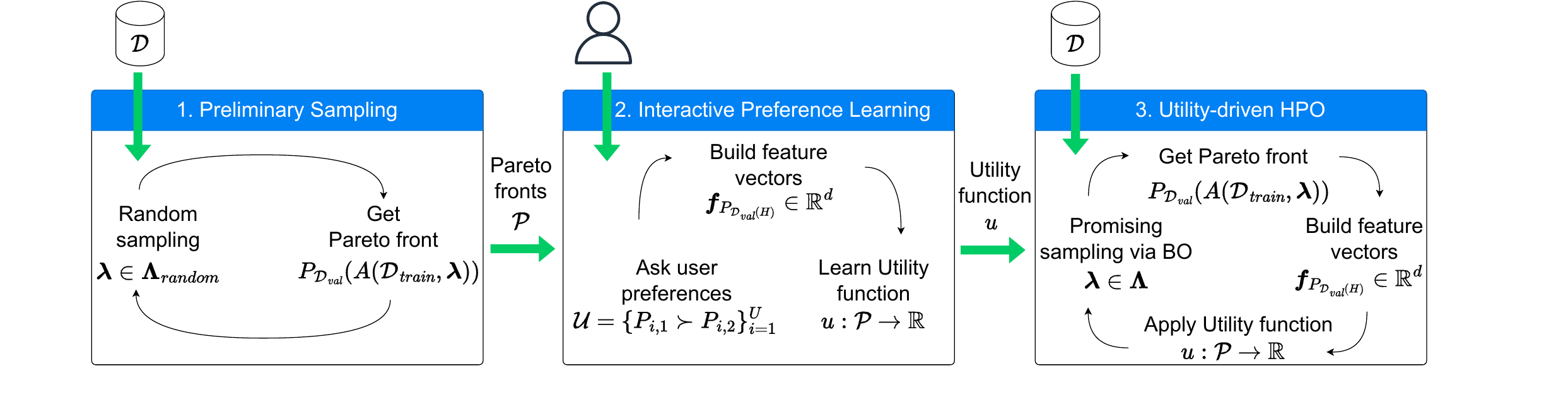} 
\caption{Overview of the three phases of our approach: Preliminary Sampling provides the user with different Pareto fronts, Interactive Preference Learning allows the user to express their preferences, finally, Utility-driven HPO guides the optimization to the user desiderata.}
\label{fig:method}
\end{figure*}

\section*{Interactive Hyperparameter Optimization in Multi-Objective Problems}
\label{sec:method}

Our approach tackles the HPO problem (cf. Eq.~\ref{eq:hpo}) for MO-ML algorithms, i.e., learning algorithms that return a Pareto front of models instead of a single model as a result of the learning process, works in three phases as depicted in Fig.~\ref{fig:method}: 
\begin{enumerate}
    \item \textbf{Preliminary Sampling}: In the preliminary sampling phase we sample a fixed but small amount of hyperparameter configurations, evaluate these configurations and store the corresponding Pareto fronts of models returned by the MO-ML algorithm. 
    \item \textbf{Interactive Preference Learning}: In the interactive preference learning phase, we construct pairs of Pareto fronts from the ones obtained in the preliminary sampling phase and show these pairs to the user, who rates which of two shown Pareto fronts they prefer. Based on the pairwise preferences obtained from the user and feature representations of the Pareto fronts underlying these pairwise preferences, we learn a latent utility function, which, given a Pareto front, outputs a utility score.
    \item \textbf{Utility-driven HPO}: In this HPO phase, we instantiate an HPO tool (e.g. SMAC \citep{hutter-lion11a,lindauer-jmlr22a} as in our experiments) with the previously learned utility function as a loss function and perform standard HPO.
\end{enumerate}

\subsection*{Preliminary Sampling}
\label{ssec:sampling}

The underlying goal of the preliminary sampling phase is to obtain a set of Pareto fronts which we can use to construct pairs of Pareto fronts, which the user can rate in the subsequent stage. Keeping in mind, that we want to learn a utility function from the pairwise comparisons provided by the user, our set of Pareto fronts should ideally reasonably cover the space of possible Pareto fronts. This avoids potential generalization problems of the learned utility function, if it is provided with a Pareto front from a part of the Pareto front space which is very far off from any training data. 

Recall that we perform HPO and our MO-ML algorithm $A: \mathbb{D} \times \vec{\Lambda} \rightarrow 2^\mathcal{H}$ returns different Pareto fronts of models for a given dataset $\mathcal{D}_\mathit{train} \in \mathbb{D}$ based on different hyperparameter configurations $\vec{\lambda} \in \vec{\Lambda}$. As such, we can obtain a set of Pareto fronts of models by sampling a fixed number of hyperparameter configurations $\vec{\Lambda}_\mathit{random} \subset \vec{\Lambda}$ at random, training the algorithm instantiated with the corresponding hyperparameter configuration on the training data $\mathcal{D}_\mathit{train}$ and evaluating it according to our loss functions $\mathcal{L}_m$ on $\mathcal{D}_\mathit{val}$. This leads to a set of Pareto fronts defined as
\begin{equation}
    \mathcal{P} = \{ P_{\mathcal{D}_\mathit{val}}(A(\mathcal{D}_\mathit{train}, \vec{\lambda})) \vert \vec{\lambda} \in \vec{\Lambda}_\mathit{random}\} \, .
\end{equation}

As the rich literature on AutoML and HPO shows, random search leads to a good coverage of the hyperparameter configuration space \citep{bischl-dmkd23a}. However, this does not automatically entail a reasonable coverage of the corresponding Pareto front space. 

\subsection*{Interactive Preference Learning}
\label{ssec:preference_learning_method}

The goal underlying this phase of our approach is construct pairs of Pareto fronts to show to the user and learn a utility function of the form $u: \mathcal{P} \rightarrow \mathbb{R}$, which, given a Pareto front, returns a utility score of the Pareto front based on the preferences obtained from the user. 

\subsubsection*{Acquiring User Preferences}

More formally, we start by constructing a set of pairs of Pareto fronts as 
\begin{equation}
    \Omega = \{ (P_1, P_2) \vert P_1,P_2 \in \mathcal{P}, P_1 \neq P_2\} \, ,
\end{equation}
which leads to a number of pairs quadratic in the number of samples Pareto fronts. Since we do not want to overwhelm the user by showing them too many of such pairs, we can fallback to subsampling $\Omega$ to decrease the number of pairs to show to the user. However, as we show in the experimental evaluation later, we can also simply have a rather short preliminary sampling phase leading to a small number of Pareto fronts and thus also to a reasonably sized set $\Omega$ of pairs of Pareto fronts without compromising too much of the performance of our approach.
Once the user has rated for each of such pairs $(P_1, P_2) \in \Omega$ whether he prefers the Pareto front $P_1$ over $P_2$, we have an object ranking dataset of the form
\begin{equation}\label{eq:object_ranking_dataset}
    \mathcal{U} = \{ P_{i,1} \succ P_{i,2}\}_{i=1}^U \, ,
\end{equation} where, without loss of generality, we assume that the user prefers Pareto front $P_{i,1}$ over Pareto front $P_{i,2}$. 

\subsubsection*{Feature Representation of Pareto Fronts}
\begin{figure}
\centering
\includegraphics[width=1.1\columnwidth]{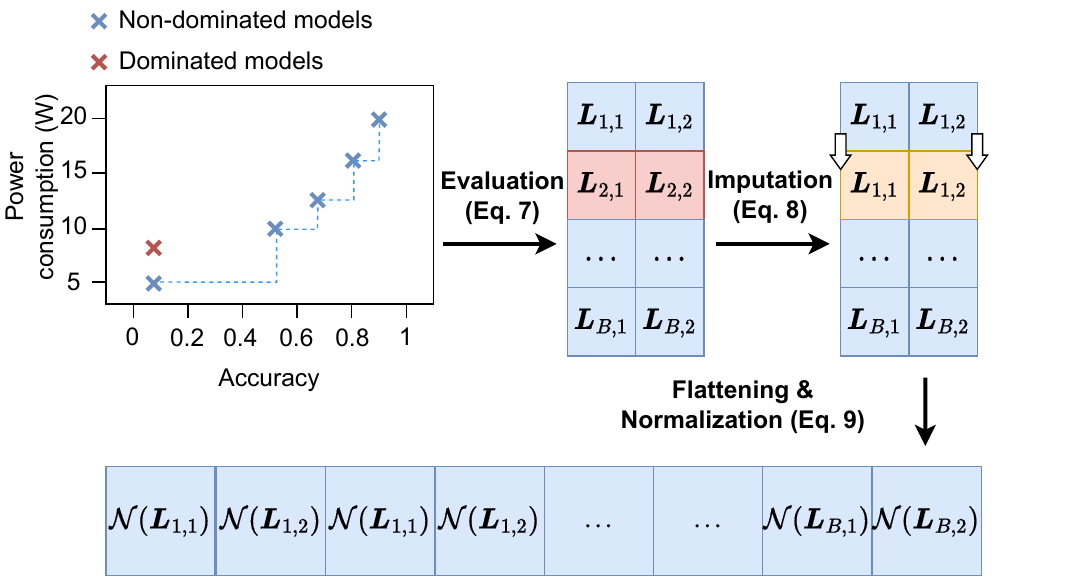} 
\caption{Visualization of the feature representation of a Pareto front based on two loss functions.}
\label{fig:preference_preparation}
\end{figure}

To learn a utility function based on the object ranking dataset defined in Eq.~\ref{eq:object_ranking_dataset}, we additionally require a feature representation of a Pareto front such that the object ranking learning algorithm we employ can generalize over unseen Pareto fronts.
More precisely, we aim to encode the Pareto front $P_{\mathcal{D}_\mathit{val}(H)}$ returned by $A$ in a $d$-dimensional feature representation $\vec{f}_{P_{\mathcal{D}_\mathit{val}(H)}} \in \mathbb{R}^d$ as depicted in Fig.~\ref{fig:preference_preparation}. To this end, we assume to have access to all models returned by $A$, i.e. $H$ including all dominated models internally learned by $A$ and that there is some order imposed on the model space $\mathcal{H}$ and hence on $H$. We evaluate all of these models $h_b \in H$ with each loss function $\mathcal{L}_i$ on $\mathcal{D}_\mathit{val}$ resulting in the following matrix
\begin{equation}
    \vec{L} = \left( \begin{array}{ccc}
         \mathcal{L}_1(h_1,\mathcal{D}_{\mathit{val}}) & \ldots & \mathcal{L}_M(h_1,\mathcal{D}_{\mathit{val}}) \\
         \vdots & \ddots & \vdots \\
         \mathcal{L}_1(h_B,\mathcal{D}_{\mathit{val}}) & \ldots & \mathcal{L}_M(h_B,\mathcal{D}_{\mathit{val}}) \\
    \end{array} \right) \, ,
\end{equation} 
where $B \geq \vert H \vert$ is the maximal number of models returned by $A$. If $A$ returns $B' < B$ values, we forward impute missing values for $B' + 1 \leq b \leq B' + (B - B')$ as 
\begin{equation}\label{eq:replace_dominated}
    \vec{L}_{m,b} \leftarrow \mathcal{L}_m(h_{b-1},\mathcal{D}_{\mathit{val}}) ~ \forall 1 \leq m \leq M \, . 
\end{equation} 
For each model $h_b$ we check if it is dominated by the previous model $h_{b-1}$ as defined in Eq.~\ref{eq:pareto_def}. If it is dominated, its loss values in the matrix are replaced by the ones from the previous non-dominated model following Eq.~\ref{eq:replace_dominated}.
At the end of this process, this matrix only contains loss values of 
models contained in the Pareto front.
Last, the matrix is flattened and standardized ($\mathcal{N}(\cdot)$) across $\Omega$: 
\begin{equation}\label{eq:feature_representation}
    \vec{f}_{P_{\mathcal{D}_\mathit{val}(H)}} = \left[ \mathcal{N}(\vec{L}_{1,1}), 
    \ldots, \mathcal{N}(\vec{L}_{B,M}) \right] \, .
\end{equation}

Note that we assume the order on our models to, on one hand ensure that the replacement and imputation strategy works as described above, and on the other hand works as a positional encoding which, assuming a suitable order relation is defined, in itself contains valuable domain knowledge. For our experimental evaluation, we define an ordering based on loss values of one of the considered loss functions.

\subsubsection*{Learning A Utility Function for Pareto Fronts}

With the object ranking dataset as defined in Eq.~\ref{eq:object_ranking_dataset} and the feature representation previously described in Eq.~\ref{eq:feature_representation}, we can now learn the utlity function. To this end, we leverage the RankSVM approach by \citet{joachims2002optimizing} which generalizes a standard SVM to the case of object ranking. This approach is very appealing for our use case as it allows to easily extract the latent utility function learned as part of the object ranker, which we want to use in the subsequent stage. 

The idea underlying the (linear) RankSVM is that for every pairwise comparison $P_{1} \succ P_{2}$ in our object ranking dataset $\mathcal{U}$ we want that, without loss of generality, the hyperplane defined by the learned weight vector $\vec{w} \in \mathbb{R}^d$ separates $P_{1}$ and $P_{2}$. Formally, it should hold for every pair: 
\begin{align}
    P_{1} \succ P_{2} &\Leftrightarrow \vec{w}^\intercal\vec{f}_{P_{1}} > \vec{w}^\intercal\vec{f}_{P_{2}} \label{eq:rank_svm_1} \\ 
    &\Leftrightarrow \vec{w}^\intercal\left(\vec{f}_{P_{1}}-\vec{f}_{P_{2}}\right) > 0 \label{eq:rank_svm_2} \\ 
    &\Leftrightarrow \vec{w}^\intercal\left(\vec{f}_{P_{2}}-\vec{f}_{P_{1}}\right) < 0 \label{eq:rank_svm_3} \, . 
\end{align}

As in the standard case of SVMs this problem is NP-hard as noted by \citep{joachims2002optimizing}, which can be solved by the standard problem relaxation leveraging slack variables as for normal SVMs. We refer to \citep{joachims2002optimizing} for details. 

With the observations formalized in Eq.~\ref{eq:rank_svm_1}-\ref{eq:rank_svm_3} in mind, one can implement a RankSVM with a standard classification SVM trained with a dataset of the form 
\begin{equation}
\begin{split}
    \mathcal{D}_{\mathit{SVM}} &= \left\{ \left(\left(\vec{f}_{P_{1}}-\vec{f}_{P_{2}}\right), 1\right) \middle\vert  P_{1} \succ P_{2} \in \mathcal{U} \right\} \\
    &\cup \left\{ \left(\left(\vec{f}_{P_{2}}-\vec{f}_{P_{1}}\right), 0\right) \middle\vert  P_{1} \succ P_{2} \in \mathcal{U} \right\} \, .
\end{split}
\end{equation}
We add a positive example to encourage Eq.~\ref{eq:rank_svm_2} and at the same time a negative example to encourage Eq.~\ref{eq:rank_svm_3}, which both enforce a balanced dataset.

After training a standard classification SVM on $\mathcal{D}_{\mathit{SVM}}$, we define the utility function $u: \mathcal{P} \rightarrow \mathbb{R}$ via the SVM weight vector $\vec{w}$ leveraging the feature representation of a Pareto front $P \in \mathcal{P}$ defined in Eq.~\ref{eq:feature_representation} as $u(P) = \vec{w}^T\vec{f}_P$.

Although we only explained the linear RankSVM, the kernel trick~\citep{scholkopf-book02} can be applied as usual in the case of SVMs leading to non-linear versions.

All in all, preference learning and in particular object ranking offers a way to quantify the quality of a Pareto front in terms of a single scalar value w.r.t. the user desiredata without asking the user for this concrete scalar value.
The RankSVM idea leverages the robustness and effectiveness of support vector machines to operationalize this idea while offering a clear and interpretable ranking mechanism. 

\subsection*{Utility-driven HPO}
\label{ssec:utility_automl}

As we learned the utility function $u: \mathcal{P} \rightarrow \mathbb{R}$ from the user preferences, we want to leverage it as a Pareto front quality indicator to optimize with a standard HPO tool. To this end, we leverage the well-known HPO tool SMAC \citep{hutter-lion11a,lindauer-jmlr22a} instantiated with the learned utility function $u$ as a cost function. 
\section*{Evaluation}
\label{sec:evaluation}

We evaluate our approach in a case-study related to decreasing the energy consumption of ML models and in particular consider an MO-ML algorithm optimizing for both accuracy and energy consumption showing how our approach can be used in the context of Green AutoML \citep{tornede-jair23a}. In the following, we first detail the general experimental setup and then present two experiments performed.

All code including detailed documentation can be found on Github\footnote{\url{https://github.com/automl/interactive-mo-ml}}, the technical appendix at the end of the document---after the references.

\subsection*{General Experimental Setup}
\label{ssec:experimental_setup}

The MO-ML algorithm, whose hyperparameters we want to tune, is a wrapper for a funnel-shaped deep neural network (DNN) learning algorithm that exposes the same hyperparameters as the underlying DNN learning algorithm, but performs a grid-search over the number of epochs to train the DNN to return a Pareto front of models (cf. Appendix D.1).

These models are all identical, but trained for different number of epochs and can thus be seen as snapshots of the DNN learning curve~\citep{mohr-arxiv22a} at different epochs. A lower number of epochs usually leads to a lower energy consumption, but also to a lower accuracy indicating that the two objectives are potentially conflicting. The concrete hyperparameters our DNN algorithms exposes are defined by LCBench \citep{zimmer-tpami21a}, a well-known multi-fidelity deep learning benchmark, and given in Table 1
in the appendix. LCBench comprises evaluations of over 2000 funnel-shaped MLP neural networks with varying hyperparameters on 35 datasets of the OpenML CC-18 suite \citep{bischl-arxiv17a}. In the spirit of Green AutoML \citep{tornede-jair23a}, we leverage the benchmark surrogate for LCBench provided by YAHPO-GYM \citep{pfisterer-arxiv21a}.\footnote{We use the surrogate model to estimate the performance of those configurations for which no result is available in LCBench.} We measure the accuracy of a model as the validation accuracy on $33\%$ of the corresponding OpenML CC-18 dataset. We estimate the power consumption in watts~($W$) for training a model by assuming the maximum consumption for the whole provided training time. The evaluations present in LCBench were performend on an Intel Xeon Gold 6242 with a maximum consumption of $150~Wh$.

As noted earlier, many users struggle to choose a Pareto front quality indicator as a loss function for an HPO tool that yields Pareto fronts with their desired shape. Our evaluation focuses on users that are likely to make a wrong decision wrt. choosing the correct quality indicator, but can very well label pariwise comparisons of Pareto fronts according to an indicator without knowing the indicator itself. As such, we simulate users by labeling pairwise comparisons according to hypervolume (HV), spacing (SP), maximum spread (MS) and R2 as Pareto front quality indicators. HV quantifies the volume of the front by merging the hypercubes determined by each of its models $h \in H$ and a reference point $r$ (i.e., the worst possible model). HV values range from $0$ to $1$, where $1$ is the optimal value. SP is one of the most popular uniformity indicators and  gauges the variation of the distance between models in Pareto front. MS is a widely used spread indicator, which measures the range of a Pareto front by considering the maximum extent on each objective. R2 measures the proximity to a specific reference point $r$ via the Chebyshev norm
(formal definition in Appendix B).

The order over the models returned by $A$, which we assume for the feature representation, 
is given by the energy consumption loss function, which means that models are ordered based on their energy consumption.


\subsection*{Experiment: Object Ranking Performance}

\begin{figure}[t]
\centering
\includegraphics[width=0.9\columnwidth]{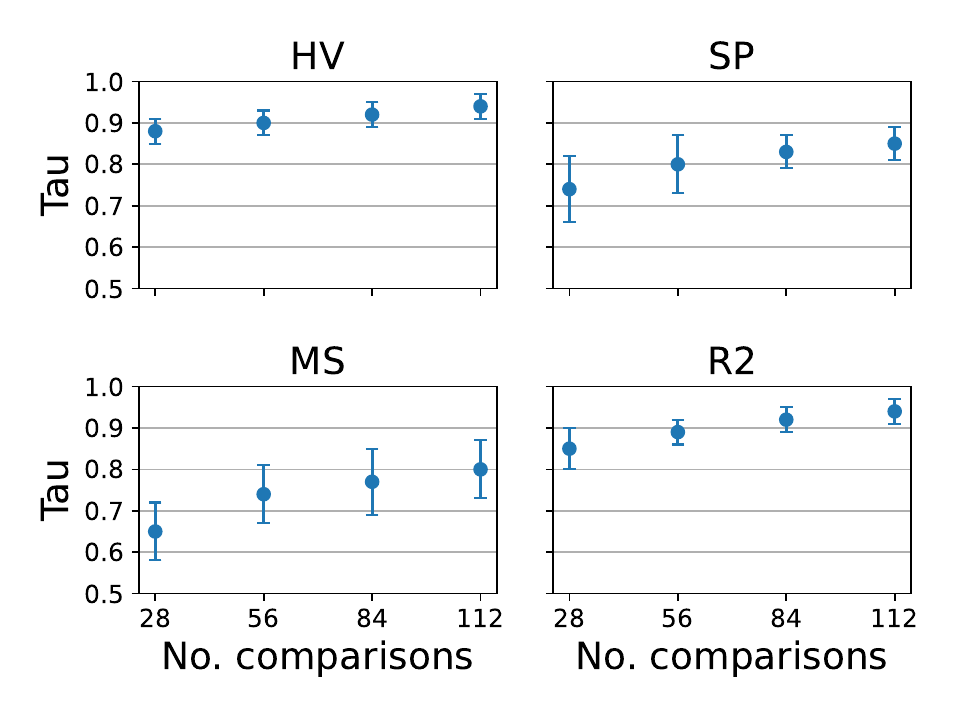} 
\caption{Kendall's Tau of the preference learning models.}
\label{fig:preference_evaluation}
\end{figure}

In the following, we evaluate our object ranker.

\paragraph{Additional Experimental Setup}
We tune the hyperparameters of our preference learning models, one for each quality indicator, on top of the first 3 datasets of LCBench: KDDCup09\_appetency, covertype, Amazon\_employee\_access.
Each configuration has been evaluated by averaging the performance achieved in those datasets over $3$ different seeds.
As to the evaluation within each dataset, we performed a cross-validation: we split the sampled Pareto fronts $\mathcal{P}$ in $5$ folds and compute all possible pairwise comparisons within each fold. 
Specifically, we set the number of sampled Pareto fronts $|\mathcal{P}|$ at $40$, so that we can create $5$ folds of $8$ elements, which translates into $\binom{8}{2} = 28$ pairwise comparisons.
Concretely, at each cross-validation evaluation, we use the pairwise comparisons of the $4$ training folds to predict the global ranking of the $8$ Pareto fronts in the test fold, and compare it with the ranking of the Pareto fronts given by the quality indicator at hand. 
These rankings are compared in terms of a ranking loss function called Kendall's Tau correlation~\citep{kendall-article48}.
Roughly speaking, this measures how correlated two rankings are. A correlation score of -1 can be interpreted as an inverse correlation, a score of $0$ as no correlation and a score of $1$ as perfect correlation.

\paragraph{Result Discussion}
\begin{figure*}[t]
\centering
    \resizebox{0.81\textwidth}{!}{
    \Huge{
	\begin{tabular}{l|c|c|c|c}
	\toprule
	PB$\backslash$IB & $HV\uparrow$ & $SP\downarrow$ & $MS\uparrow$ & $R2\downarrow$ \\ \midrule
	$HV\uparrow$ & \cellcolor{red!1}{\begin{tabular}{ccc}  0.76  & \multirow{2}{*}{$\backslash$} & \textbf{0.77} \\ ($\pm$0.17) & & \textbf{($\pm$0.17) }\end{tabular}} & \cellcolor{blue!15}{\begin{tabular}{ccc} \textbf{ 0.76}  & \multirow{2}{*}{$\backslash$} & 0.52 \\ \textbf{($\pm$0.17)} & & ($\pm$0.24) \end{tabular}} & \cellcolor{blue!15}{\begin{tabular}{ccc} \textbf{ 0.76}  & \multirow{2}{*}{$\backslash$} & 0.52 \\ \textbf{($\pm$0.17)} & & ($\pm$0.21) \end{tabular}} & \cellcolor{red!1}{\begin{tabular}{ccc}  0.76  & \multirow{2}{*}{$\backslash$} & \textbf{0.77} \\ ($\pm$0.17) & & \textbf{($\pm$0.16) }\end{tabular}} \\ \midrule
	$SP\downarrow$ & \cellcolor{blue!67}{\begin{tabular}{ccc} \textcolor{white}{\textbf{ 0.01}}  & \multirow{2}{*}{\textcolor{white}{$\backslash$}} & \textcolor{white}{0.03} \\ \textcolor{white}{\textbf{($\pm$0.01)}} & & \textcolor{white}{($\pm$0.02)} \end{tabular}} & \cellcolor{red!0}{\begin{tabular}{ccc}  \textbf{0.01}  & \multirow{2}{*}{$\backslash$} & \textbf{0.01} \\ \textbf{($\pm$0.01)} & & \textbf{($\pm$0.0) }\end{tabular}} & \cellcolor{blue!100}{\begin{tabular}{ccc} \textcolor{white}{\textbf{ 0.01}}  & \multirow{2}{*}{\textcolor{white}{$\backslash$}} & \textcolor{white}{0.04} \\ \textcolor{white}{\textbf{($\pm$0.01)}} & & \textcolor{white}{($\pm$0.03) }\end{tabular}} & \cellcolor{blue!100}{\begin{tabular}{ccc} \textcolor{white}{\textbf{ 0.01}}  & \multirow{2}{*}{\textcolor{white}{$\backslash$}} & \textcolor{white}{0.04} \\ \textcolor{white}{\textbf{($\pm$0.01)}} & & \textcolor{white}{($\pm$0.02)} \end{tabular}} \\ \midrule
	$MS\uparrow$ & \cellcolor{blue!74}{\begin{tabular}{ccc} \textcolor{white}{\textbf{ 0.61}}  & \multirow{2}{*}{\textcolor{white}{$\backslash$}} & \textcolor{white}{0.19} \\ \textcolor{white}{\textbf{($\pm$0.09)}} & & \textcolor{white}{($\pm$0.08) }\end{tabular}} & \cellcolor{blue!74}{\begin{tabular}{ccc} \textcolor{white}{\textbf{ 0.61}}  & \multirow{2}{*}{\textcolor{white}{$\backslash$}} & \textcolor{white}{0.19} \\ \textcolor{white}{\textbf{($\pm$0.09)}} & & \textcolor{white}{($\pm$0.12) }\end{tabular}} & \cellcolor{red!6}{\begin{tabular}{ccc}  0.61  & \multirow{2}{*}{$\backslash$} & \textbf{0.65} \\ ($\pm$0.09) & & \textbf{($\pm$0.06) }\end{tabular}} & \cellcolor{blue!55}{\begin{tabular}{ccc} \textbf{ 0.61}  & \multirow{2}{*}{$\backslash$} & 0.23 \\ \textbf{($\pm$0.09)} & & ($\pm$0.11) \end{tabular}} \\ \midrule
	$R2\downarrow$ & \cellcolor{red!4}{\begin{tabular}{ccc}  0.23  & \multirow{2}{*}{$\backslash$} & \textbf{0.22} \\ ($\pm$0.16) & & \textbf{($\pm$0.16) }\end{tabular}} & \cellcolor{blue!35}{\begin{tabular}{ccc} \textbf{ 0.23}  & \multirow{2}{*}{$\backslash$} & 0.47 \\ \textbf{($\pm$0.16)} & & ($\pm$0.23) \end{tabular}} & \cellcolor{blue!32}{\begin{tabular}{ccc} \textbf{ 0.23}  & \multirow{2}{*}{$\backslash$} & 0.45 \\ \textbf{($\pm$0.16)} & & ($\pm$0.21) \end{tabular}} & \cellcolor{red!9}{\begin{tabular}{ccc}  0.23  & \multirow{2}{*}{$\backslash$} & \textbf{0.21} \\ ($\pm$0.16) & & \textbf{($\pm$0.16) }\end{tabular}} \\ \midrule
	\end{tabular}
    }
    }
\caption{Comparison between indicator-based HPO (i.e., IB, columns) and preference-based HPO (i.e., PB, rows). The preference learning model is trained using 28 pairwise comparisons.}\label{tbl:end_to_end_evaluation_28}\end{figure*}

In Fig.~\ref{fig:preference_evaluation} we plot the performance of our models in terms of the Kendall's Tau averaged across the folds with the error bars defined by the corresponding standard deviation. In particular, we show how this performance varies with the number of available pairwise comparisons, i.e. the size of the object ranking training data, highlighting that, depending on the quality indicator, we obtain a reliable ranking model for Pareto fronts with rather few training data.

Generally, as to be expected, the correlation scores increase for each utility function as the number of comparisons increases independent of the quality indicator based on which they are learned. However, both the score at the lowest number of comparisons and the degree to which it increases with a growing number of examples varies substantially between the different utility functions. This indicates that it is easier for our object ranker to learn a behavior similar to HV or R2 than to SP or MS which also coincides with the fact that HV and R2 are external quality indicators whereas SP and MS are internal indicators.
A potential reason for these differences in modeling performance of our object ranker might be grounded in the feature representation of Pareto fronts we chose and in the linearity of the RankSVM used as an object ranker. In particular, both SP and MS require computations over pairs of models such that a quadratic kernel or feature transformation might be better suited for these indicators. Nevertheless, as we see in the next experiment, the ranking performance seems to be sufficient to guide the HPO process.

\subsection*{Experiment: HPO Approach Performance}
\label{sssec:end_to_end}

In the following, we leverage the remaining 32 datasets of LCBench to evaluate our complete approach from phases 1 to 3. We will demonstrate that our HPO approach performs much better than SMAC assuming a user that chooses the wrong indicator and that it performs comparable assuming an advanced user knowing which indicator to pick.

\paragraph{Additional Experimental Setup}

In order to quantify how well our approach works, we compare how well SMAC instantiated with each of the Pareto front indicators above as a loss function (IB) works compared to our approach with the user simulated as mentioned above (PB). In particular, for each Pareto front indicator, we run our approach with a simulated user that behaves according to that indicator and compare against the HPO tool instantiated with each of the indicators as a loss function. That way, we cannot only quantify how much better our approach works wrt. to each of the quality indicators, under the assumption that a user chooses a wrong quality indicator, but also that our approach does not perform substantially worse in cases where the user picks the correct indicator. We run both IB and PB for a budget of $30$ evaluations on each of the datasets for $3$ seeds and report the mean and standard deviation over the seeds and datasets.

\paragraph{Result Discussion}
Figure~\ref{tbl:end_to_end_evaluation_28} visualizes the comparison of SMAC initialized with different Pareto front quality indicators as a loss function (IB) and our approach based on the learned utility function as a Pareto front quality indicator (PB). 
The indicators in the rows represent the ones used for labeling the user preferences, and hence, to train the preference learning models.
The indicators in the columns represent the quality indicators chosen for optimization in SMAC. 
As a consequence, in each cell, we find the performance (averaged over seeds and datasets) and the respective standard deviation of the preference-based SMAC (PB) and the indicator-based SMAC (IB).
This is expressed in terms of the quality indicator leveraged to rank the Pareto front (i.e. the one given in the row), hence, providing us with an estimation of how compliant the final Pareto front is with the user preferences.
The cells in the diagonal correspond to situations where our approach is compared to IB initialized with the "correct" quality indicator function, whereas the off-diagonal cells correspond to scenarios where a user chooses the "wrong" quality indicato for IB. For a better visual interpretability, we color cells with a blue tone depending on how much better our approach (PB) is relative to IB given in the respective column and red in cases where we are worse. Moreover, we highlight the better performance value for each cell in bold.
If our learned utility function perfectly resembled the ground truth quality indicator, we would expect that the two values in each diagonal cell are identical and the coloring would be white as a consequence. Moreover, the better our approach is in one of the settings, the darker is the blue of the corresponding cell. At the same time, the worse our approach is in one of the settings, the darker will be the red in the corresponding cell. 

As the table shows, our approach (PB) behaves comparable to IB in cases where a user picks the correct Pareto front quality indicator (diagonal) and almost always better in cases where a user picks the wrong Pareto front quality indicator. In particular, we perform better or equal in $11/16$ cases whereas IB performs slightly better only in $5/10$ cases. In cases where our approach performs better, the improvements are often substantial, whereas in cases of degradation our approach is often only slightly worse.

Overall, our evaluation demonstrates that our approach successfully frees users from selecting the correct quality indicator aligned with their desiderata at the slight cost of visually comparing a low number of Paretos upfront.

\section*{Conclusion}
\label{sec:conclusion}
In this paper, we propose a human-centered interactive HPO approach tailored towards MO ML leveraging preference learning to extract desiderata from users that guide the optimization. In particular, we learn a utility function for Parto fronts based on pairwise preferences given by the user which we use as a loss function in a subsequent standard HPO process. In an experimental study, we  demonstrate that our end-to-end approach performs much better than off-the-shelf HPO with a user that chooses an indicator not aligned with their desiderata and that it performs comparable to off-the-shelf HPO operated by an advanced user knowing which indicator to pick. As such, our approach successfully frees users from selecting the correct Pareto front quality indicator aligned with their desiderata at the slight cost of visually comparing a low number of Pareto fronts upfront.

As described in more detail in the limitations section in Appendix A,
our approach naturally also offers room for future work. For example, we deem it interesting to design other feature representations with less assumptions and to generalize our approach to a larger number of loss functions as it is currently practically limited to two as users might have a hard time to rate higher dimensional Pareto fronts in the interactive part.

\section*{Acknowledgements}
Alexander Tornede and Marius Lindauer acknowledge funding by the European Union (ERC, ``ixAutoML'', grant no.101041029). Views and opinions expressed are however those of the author(s) only and do not necessarily reflect those of the European Union or the European Research Council Executive Agency. Neither the European Union nor the granting authority can be held responsible for them. 
Tanja Tornede was supported by the German Federal Ministry of the Environment, Nature Conservation, Nuclear Safety and Consumer Protection (GreenAutoML4FAS project no. 67KI32007A). 

\bibliography{shortstrings, lib, aaai24_cleaned, shortproc}

\clearpage
\newpage
\appendix
\onecolumn

\section{Limitations}\label{appendix:limitations}
Naturally, our approach suffers from limitations, which we describe below together with potential remedies. 

First, as we rely on showing Pareto fronts to the user, we make the implicit assumption of optimizing the hyperparameters of MO-ML algorithms which optimize only two loss functions $\mathcal{L}_1$ and $\mathcal{L}_2$. Although our approach would in principle also work with more loss functions, obtaining the corresponding comparison results from users would be much more complicated as showing a user a $3$-dimensional, let alone more dimensional, Pareto fronts is infeasible. A possible remedy to this problem would be to deploy dimensionality reduction techniques for multi-objectives settings~\citep{deb-kangal05} such as PCA~\citep{jolliffe-ftrc16}, LDA~\citep{balakrishnama-article98} mapping higher dimensional Pareto fronts to two dimensions for showing them to the user. 

Second, the feature representation as described in Eq.~\ref{eq:feature_representation} could be improved in general to be independent of the number of loss functions. This would simplify the application of the subsequent object ranking algorithm as the replacement and imputation strategy might be discarded. On the downside, depending on the feature representation, the positional encoding, which we assume to be of importance for the learner, might get lost.

Third, 
we note that -- ideally -- the sampled Pareto fronts used to generate pairwise comparisons with the help of the user should be sampled such that the space of possible Pareto fronts is covered to avoid generalization issues of the learned utility function. However, our current approach does not actively account for this. In fact, achieving such a coverage is a challenging task as we cannot directly sample the corresponding Pareto fronts, but instead rely on sampling hyperparamter configurations leading to the Pareto front. A potential remedy for this problem could be to meta-learn a model mapping from a hyperparameter $\vec{\lambda}$ configuration to the feature vector $\vec{f}_{P_{\mathcal{D}_{\mathit{val}}}(A(\mathcal{D}_\mathit{train}, \vec{\lambda}))}$ representing the Pareto front obtained by running $A$ with $\lambda$. If we had such a model with a quick evaluation time, we could leverage rejection sampling and sample until we have a sufficient coverage in the feature space.

Lastly, our approach is tailored towards MO-ML. However, single objective ML is much more common in practice, limiting the applicability of our approach. As such, a generalization of our approach to doing MO-HPO of single objective ML algorithms is desirable.

\section{Formal Definition of Pareto Front Quality Indicators}\label{appendix:pareto-front-quality-indicators}
\begin{description}
     \item[HV:] Hypervolume, it quantifies the volume of the front by merging the hypercubes determined by each of its solutions $h \in H$ and the reference point $r$ (i.e., the worst possible solution), formally:
     $$HV(H) = \beta \left(\bigcup_{h \in H} \{ x \mid h \prec x \prec r \} \right)$$
     where $\beta$ denotes the Lebesgue measure. This ranges from $0$ to $1$, the highest the better.
     \item[SP:] Spacing, it is the most popular uniformity indicator and it gauges the variation of the distance between solutions in a set;
     $$SP(H) = \sqrt{\frac{1}{N-1} \sum_{h \in H} ( \bar{d} - d_1 ( h, H/h ) )^2}$$
     where $\bar{d}$ is the mean of all the $d_1 ( h, H/h ) : h \in H$ and $d_1 ( h, H/h )$ means the $L^1$ norm distance (Manhattan distance) of $h$ to the set $H/h$. A SP value of zero indicates all members of the solution set are spaced equidistantly on the basis of Manhattan distance.
     \item[MS:] Maximum Spread, it is a widely used spread indicator, and it measures the range of a solution set by considering the maximum extent on each objective:
     $$MS(H) = \sqrt{ \sum_{j = 1}^{m} \max_{h, h' \in H} (\mathcal{L}_{j}(h) - \mathcal{L}_{j}(h'))^2}$$
     where $m$ denotes the number of objectives. MS is to be maximized; the higher the value, the better extensity to be claimed.
     \item[R2:] this indicator measures the proximity to a specific reference point $r$ via the Chebyshev norm:
     $$R2(H) = \min_{h \in H} d_{\inf}(h, r)$$
     Given $m$ objectives, $d_{\inf}(h, r)$ corresponds to $\max_{i=1}^{m}(\mid \mathcal{L}_i(h) - \mathcal{L}_i(r))$. Lower values translate into Pareto fronts closer to the reference point, taken as the ideal solution.
\end{description}

\section{Fisher-Jenks Algorithm}\label{appendix:fisher-jenks-algorithm}
The Fisher-Jenks algorithm addresses the problem of determining the best arrangement of ordered values into different buckets, also known as \textit{natural breaks optimization} because it effectively highlights significant transitions in the underlying distribution. We leverage this algorithm to identify similar Pareto fronts in the set of the Preliminary sampling, i.e., that should be ranked equally in a global ranking. The process starts by trying to divide the samples into two groups, evaluating all possible break points by the Jenks optimization function. This is the core of the algorithm because it drives towards the optimal buckets and consists of a trade-off between minimizing within-bucket variance and maximizing between-bucket variance. Once the optimal breakpoint is found, the data is divided into two groups based on this breakpoint. The process is then repeated recursively for each of these groups until a predetermined number of buckets is reached. By exploiting a simple elbow method, it is possible to determine the optimal number of buckets, i.e., choosing the ``knee of a curve'' as a cutoff point, where diminishing returns are no longer worth the additional cost. 
Given $n$ samples to split into $k$ buckets, the complexity of the algorithm would be $O(k \cdot n^2)$. Yet, nowadays, finding breaks for $15$ classes for a data array of $7$ million unique values now takes $20$ seconds on commodity hardware. In these experiments, we have only $40$ samples to arrange in -- approximately -- $30$ buckets, depending on the indicator.

\section{Evaluation}
In the following, we provide details about the employed MO-ML algorithm (\ref{appendix:grid-search}), implementational set-up (\ref{appendix:implementational_details}), the LCBench configuration space (\ref{appendix:lcbench}), end-to-end performance varying the number of pairwise comparisons (\ref{appendix:additional_results}).

\subsection{MO-ML Algorithm}\label{appendix:grid-search}

Since our aim is to collect Pareto fronts and get preferences on their shapes, we are interested in MO-ML algorithms that provide different Paretos as the hyperparameters change.
Fortunately, according to the case study, an extensive literature is readily available.
For instance, in the domain of fairness in AI, MO-ML implementations yield a Pareto front through mitigation, i.e., a hyper-parameter with significant impact is varied to achieve fairer outcomes.

\begin{figure}[h]
    \centering
    \includegraphics[width=0.5\columnwidth]{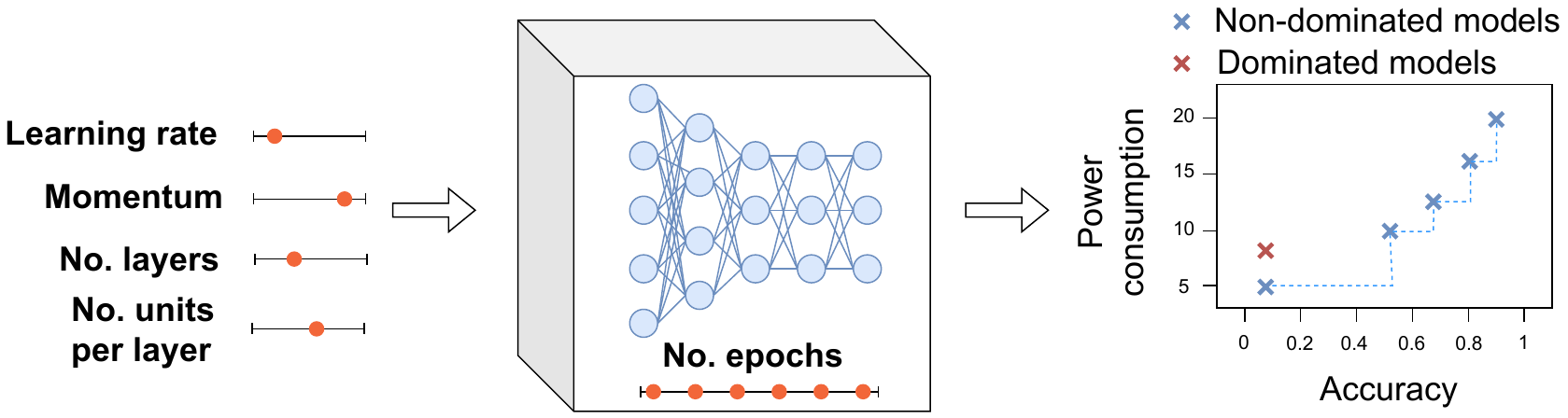} 
    \caption{Working example of the DNN wrapper: the grid search over the number of epochs allows us to have a Pareto front as output.}
    \label{fig:grid_search}
\end{figure}

In our use case, Green AutoML, Pareto fronts are handed over by observing \textit{accuracy} and \textit{power consumption} during model training of DNNs.
In practice, it is known that the \textit{number of epochs} has an obvious influence on the objectives, hence, on the shape of the retrieved Pareto front.
\Cref{fig:grid_search} shows an example: given a hyperparameter configuration, we perform a grid search over the number of epochs and obtain snapshots of the DNN learning curve~\citep{mohr-arxiv22a}.

\subsection{Implementational Details}\label{appendix:implementational_details}
The code and results of the performed experiments are available via GitHub\footnote{\url{https://github.com/automl/interactive-mo-ml}} with the corresponding documentation to run them.
There is no need for the user to locally build the project, we deployed a Docker Image that instantiates a pre-built container with all the needed Python dependencies and runs the experiments.
In particular, we leverage: the package cs-rank \citep{csrank2018, csrank2019} as a basis for our preference learning models, the package pymoo\citep{pymoo} for the quality indicator implementations, and jenkspy\footnote{\url{https://github.com/mthh/jenkspy}} for the Fisher-Jenks algorithm \citep{fisher1958homogeneity, jenks1967data}.
Finally, we exploit the well-known HPO tool SMAC \citep{hutter-lion11a,lindauer-jmlr22a} to apply Bayesian optimization.

The experiments were tested varying three different seeds on an AMD Ryzen $5$ $3600$ $6$-Core Processor with $64$ GB.
The total computation time was $21984$ seconds (around $6$ hours), for an overall power consumption of $520~W$.

\subsection{LCBench Configuration Space}\label{appendix:lcbench}
All runs feature funnel-shaped MLP nets and use SGD with cosine annealing without restarts. Overall, 7 parameters were sampled at random (4 float, 3 integer). These are reported in \Cref{tbl:lc_bench_space}. More details can be found at \url{https://github.com/automl/LCBench}.

\begin{table}[t]
\centering
\begin{tabular}{@{}llll@{}}
\toprule
Hyperparameter & Type & Range & Distribution \\ \midrule
Batch size & Integer & [16, 512] & Log \\
Learning rate & Float & [1e-4, 1e-1] & Log \\
Momentum & Float & [0.1, 0.99] & Linear \\
Weight decay & Float & [1e-5, 1e-1] & Linear \\
No. layers & Integer & [1, 5] & Linear \\
No. units per layer & Integer & [64, 1024] & Log \\
Dropout & Float & [0.0, 1.0] & Linear \\\bottomrule
\end{tabular}
\caption{Search space of LCBench.}
\label{tbl:lc_bench_space}
\end{table}

\subsection{Additional Experimental Results}\label{appendix:additional_results}
\paragraph{Robustness} In the main paper, we reported the performance of the end-to-end approach when the preference learning models are trained atop $28$ pairwise comparisons ($1$ out of the $5$ available cross-validation folds).
We also tested our approach by varying such available pairwise comparisons. Tables~\ref{tbl:end_to_end_evaluation_56} to \ref{tbl:end_to_end_evaluation_140} shows the performance when the models used from $2$ to $5$ folds, that would be -- respectively -- $56$, $84$, $112$, and $140$ pairwise comparisons.

We can observe negligible fluctuations in performance among the tables, thus the insights presented in the main paper remain consistent.
This indicates that our approach is robust; indeed, when leveraging just $28$ pairwise comparisons, we are able to get performance as good as when using $140$ comparisons.

\begin{table*}[!ht]
\centering
	\begin{tabular}{l|c|c|c|c}
	\toprule
	PB$\backslash$IB & $HV\uparrow$ & $SP\downarrow$ & $MS\uparrow$ & $R2\downarrow$ \\ \midrule
	$HV\uparrow$ & \cellcolor{red!3}{\begin{tabular}{ccc}  0.75  & \multirow{2}{*}{\Large{$\backslash$}} & \textbf{0.77} \\ \footnotesize{($\pm$0.18)} & & \footnotesize{\textbf{($\pm$0.17) }}\end{tabular}} & \cellcolor{blue!15}{\begin{tabular}{ccc} \textbf{ 0.75}  & \multirow{2}{*}{\Large{$\backslash$}} & 0.52 \\ \footnotesize{\textbf{($\pm$0.18)}} & & \footnotesize{($\pm$0.24) }\end{tabular}} & \cellcolor{blue!15}{\begin{tabular}{ccc} \textbf{ 0.75}  & \multirow{2}{*}{\Large{$\backslash$}} & 0.52 \\ \footnotesize{\textbf{($\pm$0.18)}} & & \footnotesize{($\pm$0.21) }\end{tabular}} & \cellcolor{red!3}{\begin{tabular}{ccc}  0.75  & \multirow{2}{*}{\Large{$\backslash$}} & \textbf{0.77} \\ \footnotesize{($\pm$0.18)} & & \footnotesize{\textbf{($\pm$0.16) }}\end{tabular}} \\ \midrule
	$SP\downarrow$ & \cellcolor{blue!17}{\begin{tabular}{ccc} \textbf{ 0.02}  & \multirow{2}{*}{\Large{$\backslash$}} & 0.03 \\ \footnotesize{\textbf{($\pm$0.02)}} & & \footnotesize{($\pm$0.02) }\end{tabular}} & \cellcolor{red!50}{\begin{tabular}{ccc}  0.02  & \multirow{2}{*}{\Large{$\backslash$}} & \textbf{0.01} \\ \footnotesize{($\pm$0.02)} & & \footnotesize{\textbf{($\pm$0.0) }}\end{tabular}} & \cellcolor{blue!33}{\begin{tabular}{ccc} \textbf{ 0.02}  & \multirow{2}{*}{\Large{$\backslash$}} & 0.04 \\ \footnotesize{\textbf{($\pm$0.02)}} & & \footnotesize{($\pm$0.03) }\end{tabular}} & \cellcolor{blue!33}{\begin{tabular}{ccc} \textbf{ 0.02}  & \multirow{2}{*}{\Large{$\backslash$}} & 0.04 \\ \footnotesize{\textbf{($\pm$0.02)}} & & \footnotesize{($\pm$0.02) }\end{tabular}} \\ \midrule
	$MS\uparrow$ & \cellcolor{blue!72}{\begin{tabular}{ccc} \textcolor{white}{\textbf{ 0.6}}  & \multirow{2}{*}{\textcolor{white}{\Large{$\backslash$}}} & \textcolor{white}{0.19} \\ \footnotesize{\textcolor{white}{\textbf{($\pm$0.11)}}} & & \footnotesize{\textcolor{white}{($\pm$0.08) }}\end{tabular}} & \cellcolor{blue!72}{\begin{tabular}{ccc} \textcolor{white}{\textbf{ 0.6}}  & \multirow{2}{*}{\textcolor{white}{\Large{$\backslash$}}} & \textcolor{white}{0.19} \\ \footnotesize{\textcolor{white}{\textbf{($\pm$0.11)}}} & & \footnotesize{\textcolor{white}{($\pm$0.12) }}\end{tabular}} & \cellcolor{red!8}{\begin{tabular}{ccc}  0.6  & \multirow{2}{*}{\Large{$\backslash$}} & \textbf{0.65} \\ \footnotesize{($\pm$0.11)} & & \footnotesize{\textbf{($\pm$0.06) }}\end{tabular}} & \cellcolor{blue!54}{\begin{tabular}{ccc} \textbf{ 0.6}  & \multirow{2}{*}{\Large{$\backslash$}} & 0.23 \\ \footnotesize{\textbf{($\pm$0.11)}} & & \footnotesize{($\pm$0.11) }\end{tabular}} \\ \midrule
	$R2\downarrow$ & \cellcolor{red!8}{\begin{tabular}{ccc}  0.24  & \multirow{2}{*}{\Large{$\backslash$}} & \textbf{0.22} \\ \footnotesize{($\pm$0.18)} & & \footnotesize{\textbf{($\pm$0.16) }}\end{tabular}} & \cellcolor{blue!32}{\begin{tabular}{ccc} \textbf{ 0.24}  & \multirow{2}{*}{\Large{$\backslash$}} & 0.47 \\ \footnotesize{\textbf{($\pm$0.18)}} & & \footnotesize{($\pm$0.23) }\end{tabular}} & \cellcolor{blue!29}{\begin{tabular}{ccc} \textbf{ 0.24}  & \multirow{2}{*}{\Large{$\backslash$}} & 0.45 \\ \footnotesize{\textbf{($\pm$0.18)}} & & \footnotesize{($\pm$0.21) }\end{tabular}} & \cellcolor{red!12}{\begin{tabular}{ccc}  0.24  & \multirow{2}{*}{\Large{$\backslash$}} & \textbf{0.21} \\ \footnotesize{($\pm$0.18)} & & \footnotesize{\textbf{($\pm$0.16) }}\end{tabular}} \\ \midrule
	\end{tabular}
\caption{Comparison between indicator-based HPO (i.e., IB, columns) and preference-based HPO (i.e., PB, rows). The preference learning model is trained using 56 pairwise comparisons.}\label{tbl:end_to_end_evaluation_56}\end{table*}

\begin{table*}[!ht]
\centering
	\begin{tabular}{l|c|c|c|c}
	\toprule
	PB$\backslash$IB & $HV\uparrow$ & $SP\downarrow$ & $MS\uparrow$ & $R2\downarrow$ \\ \midrule
	$HV\uparrow$ & \cellcolor{red!1}{\begin{tabular}{ccc}  0.76  & \multirow{2}{*}{\Large{$\backslash$}} & \textbf{0.77} \\ \footnotesize{($\pm$0.17)} & & \footnotesize{\textbf{($\pm$0.17) }}\end{tabular}} & \cellcolor{blue!15}{\begin{tabular}{ccc} \textbf{ 0.76}  & \multirow{2}{*}{\Large{$\backslash$}} & 0.52 \\ \footnotesize{\textbf{($\pm$0.17)}} & & \footnotesize{($\pm$0.24) }\end{tabular}} & \cellcolor{blue!15}{\begin{tabular}{ccc} \textbf{ 0.76}  & \multirow{2}{*}{\Large{$\backslash$}} & 0.52 \\ \footnotesize{\textbf{($\pm$0.17)}} & & \footnotesize{($\pm$0.21) }\end{tabular}} & \cellcolor{red!1}{\begin{tabular}{ccc}  0.76  & \multirow{2}{*}{\Large{$\backslash$}} & \textbf{0.77} \\ \footnotesize{($\pm$0.17)} & & \footnotesize{\textbf{($\pm$0.16) }}\end{tabular}} \\ \midrule
	$SP\downarrow$ & \cellcolor{blue!17}{\begin{tabular}{ccc} \textbf{ 0.02}  & \multirow{2}{*}{\Large{$\backslash$}} & 0.03 \\ \footnotesize{\textbf{($\pm$0.02)}} & & \footnotesize{($\pm$0.02) }\end{tabular}} & \cellcolor{red!50}{\begin{tabular}{ccc}  0.02  & \multirow{2}{*}{\Large{$\backslash$}} & \textbf{0.01} \\ \footnotesize{($\pm$0.02)} & & \footnotesize{\textbf{($\pm$0.0) }}\end{tabular}} & \cellcolor{blue!33}{\begin{tabular}{ccc} \textbf{ 0.02}  & \multirow{2}{*}{\Large{$\backslash$}} & 0.04 \\ \footnotesize{\textbf{($\pm$0.02)}} & & \footnotesize{($\pm$0.03) }\end{tabular}} & \cellcolor{blue!33}{\begin{tabular}{ccc} \textbf{ 0.02}  & \multirow{2}{*}{\Large{$\backslash$}} & 0.04 \\ \footnotesize{\textbf{($\pm$0.02)}} & & \footnotesize{($\pm$0.02) }\end{tabular}} \\ \midrule
	$MS\uparrow$ & \cellcolor{blue!72}{\begin{tabular}{ccc} \textcolor{white}{\textbf{ 0.6}}  & \multirow{2}{*}{\textcolor{white}{\Large{$\backslash$}}} & \textcolor{white}{0.19} \\ \footnotesize{\textcolor{white}{\textbf{($\pm$0.09)}}} & & \footnotesize{\textcolor{white}{($\pm$0.08) }}\end{tabular}} & \cellcolor{blue!72}{\begin{tabular}{ccc} \textcolor{white}{\textbf{ 0.6}}  & \multirow{2}{*}{\textcolor{white}{\Large{$\backslash$}}} & \textcolor{white}{0.19} \\ \footnotesize{\textcolor{white}{\textbf{($\pm$0.09)}}} & & \footnotesize{\textcolor{white}{($\pm$0.12) }}\end{tabular}} & \cellcolor{red!8}{\begin{tabular}{ccc}  0.6  & \multirow{2}{*}{\Large{$\backslash$}} & \textbf{0.65} \\ \footnotesize{($\pm$0.09)} & & \footnotesize{\textbf{($\pm$0.06) }}\end{tabular}} & \cellcolor{blue!54}{\begin{tabular}{ccc} \textbf{ 0.6}  & \multirow{2}{*}{\Large{$\backslash$}} & 0.23 \\ \footnotesize{\textbf{($\pm$0.09)}} & & \footnotesize{($\pm$0.11) }\end{tabular}} \\ \midrule
	$R2\downarrow$ & \cellcolor{red!0}{\begin{tabular}{ccc} \textbf{ 0.22}  & \multirow{2}{*}{\Large{$\backslash$}} & \textbf{0.22} \\ \footnotesize{\textbf{($\pm$0.17)}} & & \footnotesize{\textbf{($\pm$0.16) }}\end{tabular}} & \cellcolor{blue!38}{\begin{tabular}{ccc} \textbf{ 0.22}  & \multirow{2}{*}{\Large{$\backslash$}} & 0.47 \\ \footnotesize{\textbf{($\pm$0.17)}} & & \footnotesize{($\pm$0.23) }\end{tabular}} & \cellcolor{blue!35}{\begin{tabular}{ccc} \textbf{ 0.22}  & \multirow{2}{*}{\Large{$\backslash$}} & 0.45 \\ \footnotesize{\textbf{($\pm$0.17)}} & & \footnotesize{($\pm$0.21) }\end{tabular}} & \cellcolor{red!5}{\begin{tabular}{ccc}  0.22  & \multirow{2}{*}{\Large{$\backslash$}} & \textbf{0.21} \\ \footnotesize{($\pm$0.17)} & & \footnotesize{\textbf{($\pm$0.16) }}\end{tabular}} \\ \midrule
	\end{tabular}
\caption{Comparison between indicator-based HPO (i.e., IB, columns) and preference-based HPO (i.e., PB, rows). The preference learning model is trained using 84 pairwise comparisons.}\label{tbl:end_to_end_evaluation_84}\end{table*}

\begin{table*}[!ht]
\centering
	\begin{tabular}{l|c|c|c|c}
	\toprule
	PB$\backslash$IB & $HV\uparrow$ & $SP\downarrow$ & $MS\uparrow$ & $R2\downarrow$ \\ \midrule
	$HV\uparrow$ & \cellcolor{red!1}{\begin{tabular}{ccc}  0.76  & \multirow{2}{*}{\Large{$\backslash$}} & \textbf{0.77} \\ \footnotesize{($\pm$0.17)} & & \footnotesize{\textbf{($\pm$0.17) }}\end{tabular}} & \cellcolor{blue!15}{\begin{tabular}{ccc} \textbf{ 0.76}  & \multirow{2}{*}{\Large{$\backslash$}} & 0.52 \\ \footnotesize{\textbf{($\pm$0.17)}} & & \footnotesize{($\pm$0.24) }\end{tabular}} & \cellcolor{blue!15}{\begin{tabular}{ccc} \textbf{ 0.76}  & \multirow{2}{*}{\Large{$\backslash$}} & 0.52 \\ \footnotesize{\textbf{($\pm$0.17)}} & & \footnotesize{($\pm$0.21) }\end{tabular}} & \cellcolor{red!1}{\begin{tabular}{ccc}  0.76  & \multirow{2}{*}{\Large{$\backslash$}} & \textbf{0.77} \\ \footnotesize{($\pm$0.17)} & & \footnotesize{\textbf{($\pm$0.16) }}\end{tabular}} \\ \midrule
	$SP\downarrow$ & \cellcolor{blue!17}{\begin{tabular}{ccc} \textbf{ 0.02}  & \multirow{2}{*}{\Large{$\backslash$}} & 0.03 \\ \footnotesize{\textbf{($\pm$0.02)}} & & \footnotesize{($\pm$0.02) }\end{tabular}} & \cellcolor{red!50}{\begin{tabular}{ccc}  0.02  & \multirow{2}{*}{\Large{$\backslash$}} & \textbf{0.01} \\ \footnotesize{($\pm$0.02)} & & \footnotesize{\textbf{($\pm$0.0) }}\end{tabular}} & \cellcolor{blue!33}{\begin{tabular}{ccc} \textbf{ 0.02}  & \multirow{2}{*}{\Large{$\backslash$}} & 0.04 \\ \footnotesize{\textbf{($\pm$0.02)}} & & \footnotesize{($\pm$0.03) }\end{tabular}} & \cellcolor{blue!33}{\begin{tabular}{ccc} \textbf{ 0.02}  & \multirow{2}{*}{\Large{$\backslash$}} & 0.04 \\ \footnotesize{\textbf{($\pm$0.02)}} & & \footnotesize{($\pm$0.02) }\end{tabular}} \\ \midrule
	$MS\uparrow$ & \cellcolor{blue!72}{\begin{tabular}{ccc} \textcolor{white}{\textbf{ 0.6}}  & \multirow{2}{*}{\textcolor{white}{\Large{$\backslash$}}} & \textcolor{white}{0.19} \\ \footnotesize{\textcolor{white}{\textbf{($\pm$0.1)}}} & & \footnotesize{\textcolor{white}{($\pm$0.08) }}\end{tabular}} & \cellcolor{blue!72}{\begin{tabular}{ccc} \textcolor{white}{\textbf{ 0.6}}  & \multirow{2}{*}{\textcolor{white}{\Large{$\backslash$}}} & \textcolor{white}{0.19} \\ \footnotesize{\textcolor{white}{\textbf{($\pm$0.1)}}} & & \footnotesize{\textcolor{white}{($\pm$0.12) }}\end{tabular}} & \cellcolor{red!8}{\begin{tabular}{ccc}  0.6  & \multirow{2}{*}{\Large{$\backslash$}} & \textbf{0.65} \\ \footnotesize{($\pm$0.1)} & & \footnotesize{\textbf{($\pm$0.06) }}\end{tabular}} & \cellcolor{blue!54}{\begin{tabular}{ccc} \textbf{ 0.6}  & \multirow{2}{*}{\Large{$\backslash$}} & 0.23 \\ \footnotesize{\textbf{($\pm$0.1)}} & & \footnotesize{($\pm$0.11) }\end{tabular}} \\ \midrule
	$R2\downarrow$ & \cellcolor{red!8}{\begin{tabular}{ccc}  0.24  & \multirow{2}{*}{\Large{$\backslash$}} & \textbf{0.22} \\ \footnotesize{($\pm$0.17)} & & \footnotesize{\textbf{($\pm$0.16) }}\end{tabular}} & \cellcolor{blue!32}{\begin{tabular}{ccc} \textbf{ 0.24}  & \multirow{2}{*}{\Large{$\backslash$}} & 0.47 \\ \footnotesize{\textbf{($\pm$0.17)}} & & \footnotesize{($\pm$0.23) }\end{tabular}} & \cellcolor{blue!29}{\begin{tabular}{ccc} \textbf{ 0.24}  & \multirow{2}{*}{\Large{$\backslash$}} & 0.45 \\ \footnotesize{\textbf{($\pm$0.17)}} & & \footnotesize{($\pm$0.21) }\end{tabular}} & \cellcolor{red!12}{\begin{tabular}{ccc}  0.24  & \multirow{2}{*}{\Large{$\backslash$}} & \textbf{0.21} \\ \footnotesize{($\pm$0.17)} & & \footnotesize{\textbf{($\pm$0.16) }}\end{tabular}} \\ \midrule
	\end{tabular}
\caption{Comparison between indicator-based HPO (i.e., IB, columns) and preference-based HPO (i.e., PB, rows). The preference learning model is trained using 112 pairwise comparisons.}\label{tbl:end_to_end_evaluation_112}\end{table*}

\begin{table*}[!ht]
\centering
	\begin{tabular}{l|c|c|c|c}
	\toprule
	PB$\backslash$IB & $HV\uparrow$ & $SP\downarrow$ & $MS\uparrow$ & $R2\downarrow$ \\ \midrule
	$HV\uparrow$ & \cellcolor{red!1}{\begin{tabular}{ccc}  0.76  & \multirow{2}{*}{\Large{$\backslash$}} & \textbf{0.77} \\ \footnotesize{($\pm$0.17)} & & \footnotesize{\textbf{($\pm$0.17) }}\end{tabular}} & \cellcolor{blue!15}{\begin{tabular}{ccc} \textbf{ 0.76}  & \multirow{2}{*}{\Large{$\backslash$}} & 0.52 \\ \footnotesize{\textbf{($\pm$0.17)}} & & \footnotesize{($\pm$0.24) }\end{tabular}} & \cellcolor{blue!15}{\begin{tabular}{ccc} \textbf{ 0.76}  & \multirow{2}{*}{\Large{$\backslash$}} & 0.52 \\ \footnotesize{\textbf{($\pm$0.17)}} & & \footnotesize{($\pm$0.21) }\end{tabular}} & \cellcolor{red!1}{\begin{tabular}{ccc}  0.76  & \multirow{2}{*}{\Large{$\backslash$}} & \textbf{0.77} \\ \footnotesize{($\pm$0.17)} & & \footnotesize{\textbf{($\pm$0.16) }}\end{tabular}} \\ \midrule
	$SP\downarrow$ & \cellcolor{blue!67}{\begin{tabular}{ccc} \textcolor{white}{\textbf{ 0.01}}  & \multirow{2}{*}{\textcolor{white}{\Large{$\backslash$}}} & \textcolor{white}{0.03} \\ \footnotesize{\textcolor{white}{\textbf{($\pm$0.02)}}} & & \footnotesize{\textcolor{white}{($\pm$0.02) }}\end{tabular}} & \cellcolor{red!0}{\begin{tabular}{ccc}  \textbf{0.01}  & \multirow{2}{*}{\Large{$\backslash$}} & \textbf{0.01} \\ \footnotesize{\textbf{($\pm$0.02)}} & & \footnotesize{\textbf{($\pm$0.0) }}\end{tabular}} & \cellcolor{blue!100}{\begin{tabular}{ccc} \textcolor{white}{\textbf{ 0.01}}  & \multirow{2}{*}{\textcolor{white}{\Large{$\backslash$}}} & \textcolor{white}{0.04} \\ \footnotesize{\textcolor{white}{\textbf{($\pm$0.02)}}} & & \footnotesize{\textcolor{white}{($\pm$0.03) }}\end{tabular}} & \cellcolor{blue!100}{\begin{tabular}{ccc} \textcolor{white}{\textbf{ 0.01}}  & \multirow{2}{*}{\textcolor{white}{\Large{$\backslash$}}} & \textcolor{white}{0.04} \\ \footnotesize{\textcolor{white}{\textbf{($\pm$0.02)}}} & & \footnotesize{\textcolor{white}{($\pm$0.02) }}\end{tabular}} \\ \midrule
	$MS\uparrow$ & \cellcolor{blue!72}{\begin{tabular}{ccc} \textcolor{white}{\textbf{ 0.6}}  & \multirow{2}{*}{\textcolor{white}{\Large{$\backslash$}}} & \textcolor{white}{0.19} \\ \footnotesize{\textcolor{white}{\textbf{($\pm$0.11)}}} & & \footnotesize{\textcolor{white}{($\pm$0.08) }}\end{tabular}} & \cellcolor{blue!72}{\begin{tabular}{ccc} \textcolor{white}{\textbf{ 0.6}}  & \multirow{2}{*}{\textcolor{white}{\Large{$\backslash$}}} & \textcolor{white}{0.19} \\ \footnotesize{\textcolor{white}{\textbf{($\pm$0.11)}}} & & \footnotesize{\textcolor{white}{($\pm$0.12) }}\end{tabular}} & \cellcolor{red!8}{\begin{tabular}{ccc}  0.6  & \multirow{2}{*}{\Large{$\backslash$}} & \textbf{0.65} \\ \footnotesize{($\pm$0.11)} & & \footnotesize{\textbf{($\pm$0.06) }}\end{tabular}} & \cellcolor{blue!54}{\begin{tabular}{ccc} \textbf{ 0.6}  & \multirow{2}{*}{\Large{$\backslash$}} & 0.23 \\ \footnotesize{\textbf{($\pm$0.11)}} & & \footnotesize{($\pm$0.11) }\end{tabular}} \\ \midrule
	$R2\downarrow$ & \cellcolor{red!4}{\begin{tabular}{ccc}  0.23  & \multirow{2}{*}{\Large{$\backslash$}} & \textbf{0.22} \\ \footnotesize{($\pm$0.17)} & & \footnotesize{\textbf{($\pm$0.16) }}\end{tabular}} & \cellcolor{blue!35}{\begin{tabular}{ccc} \textbf{ 0.23}  & \multirow{2}{*}{\Large{$\backslash$}} & 0.47 \\ \footnotesize{\textbf{($\pm$0.17)}} & & \footnotesize{($\pm$0.23) }\end{tabular}} & \cellcolor{blue!32}{\begin{tabular}{ccc} \textbf{ 0.23}  & \multirow{2}{*}{\Large{$\backslash$}} & 0.45 \\ \footnotesize{\textbf{($\pm$0.17)}} & & \footnotesize{($\pm$0.21) }\end{tabular}} & \cellcolor{red!9}{\begin{tabular}{ccc}  0.23  & \multirow{2}{*}{\Large{$\backslash$}} & \textbf{0.21} \\ \footnotesize{($\pm$0.17)} & & \footnotesize{\textbf{($\pm$0.16) }}\end{tabular}} \\ \midrule
	\end{tabular}
\caption{Comparison between indicator-based HPO (i.e., IB, columns) and preference-based HPO (i.e., PB, rows). The preference learning model is trained using 140 pairwise comparisons.}\label{tbl:end_to_end_evaluation_140}\end{table*}

\paragraph{Comparison to a Multi-Objective HPO algorithm}
A multi-objective HPO algorithm will need a MO metric. The main contribution of our paper is to show how to get around choosing this MO metric explicitly by using an interactive data-driven preference-based alternative. Therefore, a direct comparison is not fair from our point of view. Nevertheless, we performed a comparison per the reviewers' request showing that we perform competitively or better (\Cref{tbl:comparison}).

\begin{table}[!ht]
\centering
	\begin{tabular}{l|c}
	\toprule
    Indicator & Us \Large{$\backslash$} ParEGO \\ \midrule
	$HV\uparrow$ &  \cellcolor{blue!1}{\begin{tabular}{ccc} \textbf{0.76}  & \multirow{2}{*}{\Large{$\backslash$}} & 0.75 \\ \footnotesize{\textbf{($\pm$0.17)}} & & \footnotesize{($\pm$0.16) }\end{tabular}} \\ \midrule
	$SP\downarrow$ & \cellcolor{blue!100}{\begin{tabular}{ccc} \textcolor{white}{\textbf{0.01}}  & \multirow{2}{*}{\textcolor{white}{\Large{$\backslash$}}} & \textcolor{white}{0.04} \\ \footnotesize{\textcolor{white}{\textbf{($\pm$0.01)}}} & & \footnotesize{\textcolor{white}{($\pm$0.04) }}\end{tabular}} \\ \midrule
	$MS\uparrow$ & \cellcolor{blue!100}{\begin{tabular}{ccc} \textcolor{white}{\textbf{0.61}}  & \multirow{2}{*}{\textcolor{white}{\Large{$\backslash$}}} & \textcolor{white}{0.20} \\ \footnotesize{\textcolor{white}{\textbf{($\pm$0.09)}}} & & \footnotesize{\textcolor{white}{($\pm$0.16) }}\end{tabular}} \\ \midrule
	$R2\downarrow$ & \cellcolor{red!5}{\begin{tabular}{ccc}  0.23  & \multirow{2}{*}{\Large{$\backslash$}} & \textbf{0.22} \\ \footnotesize{($\pm$0.16)} & & \footnotesize{\textbf{($\pm$0.16) }}\end{tabular}} \\ \midrule
	\end{tabular}
\caption{Comparison with ParEGO, performance are measured w.r.t. the indicator picked by the user.}\label{tbl:comparison}\end{table}

\begin{figure*}[t]
\centering
\includegraphics[width=0.4\columnwidth]{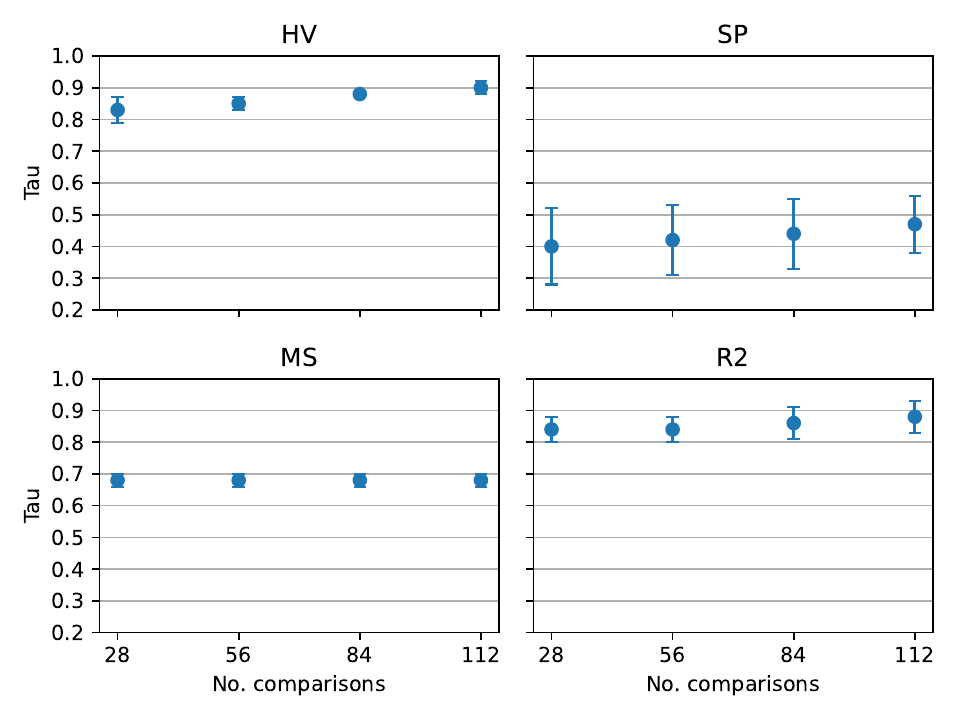} 
\caption{Kendall's Tau of the preference learning models trained with non-linear RankSVM.}
\label{fig:preference_evaluation_nonlinear}
\end{figure*}

\paragraph{Non-linear RankSVM}
We performed the end-to-end approach with a non-linear implementation of RankSVM (Chen et al. 2017) as a preference model, but cannot show the complete results here due to size constraints. Even though the ranking performance results for Kendall's Tau are worse (\Cref{fig:preference_evaluation_nonlinear}), the end-to-end performance is comparable (Tables~\ref{tbl:end_to_end_evaluation_56_nonlinear} to \ref{tbl:end_to_end_evaluation_140_nonlinear}). It seems that the larger ranking error does not influence the performance of the HPO process too much. We believe that the more powerful hypothesis class of the non-linear SVM suffers from overfitting leading to the larger ranking error. Overall, this validates our conclusion from Section 5.2.

\begin{table*}[!ht]
\centering
	\begin{tabular}{l|c|c|c|c}
	\toprule
	PB$\backslash$IB & $HV\uparrow$ & $SP\downarrow$ & $MS\uparrow$ & $R2\downarrow$ \\ \midrule
	$HV\uparrow$ & \cellcolor{red!3}{\begin{tabular}{ccc}  0.75  & \multirow{2}{*}{\Large{$\backslash$}} & \textbf{0.77} \\ \footnotesize{($\pm$0.17)} & & \footnotesize{\textbf{($\pm$0.17) }}\end{tabular}} & \cellcolor{blue!15}{\begin{tabular}{ccc} \textbf{ 0.75}  & \multirow{2}{*}{\Large{$\backslash$}} & 0.52 \\ \footnotesize{\textbf{($\pm$0.17)}} & & \footnotesize{($\pm$0.24) }\end{tabular}} & \cellcolor{blue!15}{\begin{tabular}{ccc} \textbf{ 0.75}  & \multirow{2}{*}{\Large{$\backslash$}} & 0.52 \\ \footnotesize{\textbf{($\pm$0.17)}} & & \footnotesize{($\pm$0.21) }\end{tabular}} & \cellcolor{red!3}{\begin{tabular}{ccc}  0.75  & \multirow{2}{*}{\Large{$\backslash$}} & \textbf{0.77} \\ \footnotesize{($\pm$0.17)} & & \footnotesize{\textbf{($\pm$0.16) }}\end{tabular}} \\ \midrule
	$SP\downarrow$ & \cellcolor{blue!17}{\begin{tabular}{ccc} \textbf{ 0.02}  & \multirow{2}{*}{\Large{$\backslash$}} & 0.03 \\ \footnotesize{\textbf{($\pm$0.01)}} & & \footnotesize{($\pm$0.02) }\end{tabular}} & \cellcolor{red!50}{\begin{tabular}{ccc}  0.02  & \multirow{2}{*}{\Large{$\backslash$}} & \textbf{0.01} \\ \footnotesize{($\pm$0.01)} & & \footnotesize{\textbf{($\pm$0.0) }}\end{tabular}} & \cellcolor{blue!33}{\begin{tabular}{ccc} \textbf{ 0.02}  & \multirow{2}{*}{\Large{$\backslash$}} & 0.04 \\ \footnotesize{\textbf{($\pm$0.01)}} & & \footnotesize{($\pm$0.03) }\end{tabular}} & \cellcolor{blue!33}{\begin{tabular}{ccc} \textbf{ 0.02}  & \multirow{2}{*}{\Large{$\backslash$}} & 0.04 \\ \footnotesize{\textbf{($\pm$0.01)}} & & \footnotesize{($\pm$0.02) }\end{tabular}} \\ \midrule
	$MS\uparrow$ & \cellcolor{blue!75}{\begin{tabular}{ccc} \textcolor{white}{\textbf{ 0.62}}  & \multirow{2}{*}{\textcolor{white}{\Large{$\backslash$}}} & \textcolor{white}{0.19} \\ \footnotesize{\textcolor{white}{\textbf{($\pm$0.09)}}} & & \footnotesize{\textcolor{white}{($\pm$0.08) }}\end{tabular}} & \cellcolor{blue!75}{\begin{tabular}{ccc} \textcolor{white}{\textbf{ 0.62}}  & \multirow{2}{*}{\textcolor{white}{\Large{$\backslash$}}} & \textcolor{white}{0.19} \\ \footnotesize{\textcolor{white}{\textbf{($\pm$0.09)}}} & & \footnotesize{\textcolor{white}{($\pm$0.12) }}\end{tabular}} & \cellcolor{red!5}{\begin{tabular}{ccc}  0.62  & \multirow{2}{*}{\Large{$\backslash$}} & \textbf{0.65} \\ \footnotesize{($\pm$0.09)} & & \footnotesize{\textbf{($\pm$0.06) }}\end{tabular}} & \cellcolor{blue!57}{\begin{tabular}{ccc} \textbf{ 0.62}  & \multirow{2}{*}{\Large{$\backslash$}} & 0.23 \\ \footnotesize{\textbf{($\pm$0.09)}} & & \footnotesize{($\pm$0.11) }\end{tabular}} \\ \midrule
	$R2\downarrow$ & \cellcolor{red!4}{\begin{tabular}{ccc}  0.23  & \multirow{2}{*}{\Large{$\backslash$}} & \textbf{0.22} \\ \footnotesize{($\pm$0.16)} & & \footnotesize{\textbf{($\pm$0.16) }}\end{tabular}} & \cellcolor{blue!35}{\begin{tabular}{ccc} \textbf{ 0.23}  & \multirow{2}{*}{\Large{$\backslash$}} & 0.47 \\ \footnotesize{\textbf{($\pm$0.16)}} & & \footnotesize{($\pm$0.23) }\end{tabular}} & \cellcolor{blue!32}{\begin{tabular}{ccc} \textbf{ 0.23}  & \multirow{2}{*}{\Large{$\backslash$}} & 0.45 \\ \footnotesize{\textbf{($\pm$0.16)}} & & \footnotesize{($\pm$0.21) }\end{tabular}} & \cellcolor{red!9}{\begin{tabular}{ccc}  0.23  & \multirow{2}{*}{\Large{$\backslash$}} & \textbf{0.21} \\ \footnotesize{($\pm$0.16)} & & \footnotesize{\textbf{($\pm$0.16) }}\end{tabular}} \\ \midrule
	\end{tabular}
\caption{Comparison between indicator-based HPO (i.e., IB, columns) and preference-based HPO (i.e., PB, rows). The preference learning model is trained using the non-linear RankSVM implementation and 28 pairwise comparisons.}\label{tbl:end_to_end_evaluation_28_nonlinear}\end{table*}

\begin{table*}[!ht]
\centering
	\begin{tabular}{l|c|c|c|c}
	\toprule
	PB$\backslash$IB & $HV\uparrow$ & $SP\downarrow$ & $MS\uparrow$ & $R2\downarrow$ \\ \midrule
	$HV\uparrow$ & \cellcolor{red!5}{\begin{tabular}{ccc}  0.73  & \multirow{2}{*}{\Large{$\backslash$}} & \textbf{0.77} \\ \footnotesize{($\pm$0.19)} & & \footnotesize{\textbf{($\pm$0.17) }}\end{tabular}} & \cellcolor{blue!13}{\begin{tabular}{ccc} \textbf{ 0.73}  & \multirow{2}{*}{\Large{$\backslash$}} & 0.52 \\ \footnotesize{\textbf{($\pm$0.19)}} & & \footnotesize{($\pm$0.24) }\end{tabular}} & \cellcolor{blue!13}{\begin{tabular}{ccc} \textbf{ 0.73}  & \multirow{2}{*}{\Large{$\backslash$}} & 0.52 \\ \footnotesize{\textbf{($\pm$0.19)}} & & \footnotesize{($\pm$0.21) }\end{tabular}} & \cellcolor{red!5}{\begin{tabular}{ccc}  0.73  & \multirow{2}{*}{\Large{$\backslash$}} & \textbf{0.77} \\ \footnotesize{($\pm$0.19)} & & \footnotesize{\textbf{($\pm$0.16) }}\end{tabular}} \\ \midrule
	$SP\downarrow$ & \cellcolor{blue!17}{\begin{tabular}{ccc} \textbf{ 0.02}  & \multirow{2}{*}{\Large{$\backslash$}} & 0.03 \\ \footnotesize{\textbf{($\pm$0.01)}} & & \footnotesize{($\pm$0.02) }\end{tabular}} & \cellcolor{red!50}{\begin{tabular}{ccc}  0.02  & \multirow{2}{*}{\Large{$\backslash$}} & \textbf{0.01} \\ \footnotesize{($\pm$0.01)} & & \footnotesize{\textbf{($\pm$0.0) }}\end{tabular}} & \cellcolor{blue!33}{\begin{tabular}{ccc} \textbf{ 0.02}  & \multirow{2}{*}{\Large{$\backslash$}} & 0.04 \\ \footnotesize{\textbf{($\pm$0.01)}} & & \footnotesize{($\pm$0.03) }\end{tabular}} & \cellcolor{blue!33}{\begin{tabular}{ccc} \textbf{ 0.02}  & \multirow{2}{*}{\Large{$\backslash$}} & 0.04 \\ \footnotesize{\textbf{($\pm$0.01)}} & & \footnotesize{($\pm$0.02) }\end{tabular}} \\ \midrule
	$MS\uparrow$ & \cellcolor{blue!77}{\begin{tabular}{ccc} \textcolor{white}{\textbf{ 0.63}}  & \multirow{2}{*}{\textcolor{white}{\Large{$\backslash$}}} & \textcolor{white}{0.19} \\ \footnotesize{\textcolor{white}{\textbf{($\pm$0.07)}}} & & \footnotesize{\textcolor{white}{($\pm$0.08) }}\end{tabular}} & \cellcolor{blue!77}{\begin{tabular}{ccc} \textcolor{white}{\textbf{ 0.63}}  & \multirow{2}{*}{\textcolor{white}{\Large{$\backslash$}}} & \textcolor{white}{0.19} \\ \footnotesize{\textcolor{white}{\textbf{($\pm$0.07)}}} & & \footnotesize{\textcolor{white}{($\pm$0.12) }}\end{tabular}} & \cellcolor{red!3}{\begin{tabular}{ccc}  0.63  & \multirow{2}{*}{\Large{$\backslash$}} & \textbf{0.65} \\ \footnotesize{($\pm$0.07)} & & \footnotesize{\textbf{($\pm$0.06) }}\end{tabular}} & \cellcolor{blue!58}{\begin{tabular}{ccc} \textbf{ 0.63}  & \multirow{2}{*}{\Large{$\backslash$}} & 0.23 \\ \footnotesize{\textbf{($\pm$0.07)}} & & \footnotesize{($\pm$0.11) }\end{tabular}} \\ \midrule
	$R2\downarrow$ & \cellcolor{red!4}{\begin{tabular}{ccc}  0.23  & \multirow{2}{*}{\Large{$\backslash$}} & \textbf{0.22} \\ \footnotesize{($\pm$0.17)} & & \footnotesize{\textbf{($\pm$0.16) }}\end{tabular}} & \cellcolor{blue!35}{\begin{tabular}{ccc} \textbf{ 0.23}  & \multirow{2}{*}{\Large{$\backslash$}} & 0.47 \\ \footnotesize{\textbf{($\pm$0.17)}} & & \footnotesize{($\pm$0.23) }\end{tabular}} & \cellcolor{blue!32}{\begin{tabular}{ccc} \textbf{ 0.23}  & \multirow{2}{*}{\Large{$\backslash$}} & 0.45 \\ \footnotesize{\textbf{($\pm$0.17)}} & & \footnotesize{($\pm$0.21) }\end{tabular}} & \cellcolor{red!9}{\begin{tabular}{ccc}  0.23  & \multirow{2}{*}{\Large{$\backslash$}} & \textbf{0.21} \\ \footnotesize{($\pm$0.17)} & & \footnotesize{\textbf{($\pm$0.16) }}\end{tabular}} \\ \midrule
	\end{tabular}
\caption{Comparison between indicator-based HPO (i.e., IB, columns) and preference-based HPO (i.e., PB, rows). The preference learning model is trained using the non-linear RankSVM implementation and 56 pairwise comparisons.}\label{tbl:end_to_end_evaluation_56_nonlinear}\end{table*}

\begin{table*}[!ht]
\centering
	\begin{tabular}{l|c|c|c|c}
	\toprule
	PB$\backslash$IB & $HV\uparrow$ & $SP\downarrow$ & $MS\uparrow$ & $R2\downarrow$ \\ \midrule
	$HV\uparrow$ & \cellcolor{red!1}{\begin{tabular}{ccc}  0.76  & \multirow{2}{*}{\Large{$\backslash$}} & \textbf{0.77} \\ \footnotesize{($\pm$0.16)} & & \footnotesize{\textbf{($\pm$0.17) }}\end{tabular}} & \cellcolor{blue!15}{\begin{tabular}{ccc} \textbf{ 0.76}  & \multirow{2}{*}{\Large{$\backslash$}} & 0.52 \\ \footnotesize{\textbf{($\pm$0.16)}} & & \footnotesize{($\pm$0.24) }\end{tabular}} & \cellcolor{blue!15}{\begin{tabular}{ccc} \textbf{ 0.76}  & \multirow{2}{*}{\Large{$\backslash$}} & 0.52 \\ \footnotesize{\textbf{($\pm$0.16)}} & & \footnotesize{($\pm$0.21) }\end{tabular}} & \cellcolor{red!1}{\begin{tabular}{ccc}  0.76  & \multirow{2}{*}{\Large{$\backslash$}} & \textbf{0.77} \\ \footnotesize{($\pm$0.16)} & & \footnotesize{\textbf{($\pm$0.16) }}\end{tabular}} \\ \midrule
	$SP\downarrow$ & \cellcolor{blue!17}{\begin{tabular}{ccc} \textbf{ 0.02}  & \multirow{2}{*}{\Large{$\backslash$}} & 0.03 \\ \footnotesize{\textbf{($\pm$0.01)}} & & \footnotesize{($\pm$0.02) }\end{tabular}} & \cellcolor{red!50}{\begin{tabular}{ccc}  0.02  & \multirow{2}{*}{\Large{$\backslash$}} & \textbf{0.01} \\ \footnotesize{($\pm$0.01)} & & \footnotesize{\textbf{($\pm$0.0) }}\end{tabular}} & \cellcolor{blue!33}{\begin{tabular}{ccc} \textbf{ 0.02}  & \multirow{2}{*}{\Large{$\backslash$}} & 0.04 \\ \footnotesize{\textbf{($\pm$0.01)}} & & \footnotesize{($\pm$0.03) }\end{tabular}} & \cellcolor{blue!33}{\begin{tabular}{ccc} \textbf{ 0.02}  & \multirow{2}{*}{\Large{$\backslash$}} & 0.04 \\ \footnotesize{\textbf{($\pm$0.01)}} & & \footnotesize{($\pm$0.02) }\end{tabular}} \\ \midrule
	$MS\uparrow$ & \cellcolor{blue!74}{\begin{tabular}{ccc} \textcolor{white}{\textbf{ 0.61}}  & \multirow{2}{*}{\textcolor{white}{\Large{$\backslash$}}} & \textcolor{white}{0.19} \\ \footnotesize{\textcolor{white}{\textbf{($\pm$0.1)}}} & & \footnotesize{\textcolor{white}{($\pm$0.08) }}\end{tabular}} & \cellcolor{blue!74}{\begin{tabular}{ccc} \textcolor{white}{\textbf{ 0.61}}  & \multirow{2}{*}{\textcolor{white}{\Large{$\backslash$}}} & \textcolor{white}{0.19} \\ \footnotesize{\textcolor{white}{\textbf{($\pm$0.1)}}} & & \footnotesize{\textcolor{white}{($\pm$0.12) }}\end{tabular}} & \cellcolor{red!6}{\begin{tabular}{ccc}  0.61  & \multirow{2}{*}{\Large{$\backslash$}} & \textbf{0.65} \\ \footnotesize{($\pm$0.1)} & & \footnotesize{\textbf{($\pm$0.06) }}\end{tabular}} & \cellcolor{blue!55}{\begin{tabular}{ccc} \textbf{ 0.61}  & \multirow{2}{*}{\Large{$\backslash$}} & 0.23 \\ \footnotesize{\textbf{($\pm$0.1)}} & & \footnotesize{($\pm$0.11) }\end{tabular}} \\ \midrule
	$R2\downarrow$ & \cellcolor{red!4}{\begin{tabular}{ccc}  0.23  & \multirow{2}{*}{\Large{$\backslash$}} & \textbf{0.22} \\ \footnotesize{($\pm$0.16)} & & \footnotesize{\textbf{($\pm$0.16) }}\end{tabular}} & \cellcolor{blue!35}{\begin{tabular}{ccc} \textbf{ 0.23}  & \multirow{2}{*}{\Large{$\backslash$}} & 0.47 \\ \footnotesize{\textbf{($\pm$0.16)}} & & \footnotesize{($\pm$0.23) }\end{tabular}} & \cellcolor{blue!32}{\begin{tabular}{ccc} \textbf{ 0.23}  & \multirow{2}{*}{\Large{$\backslash$}} & 0.45 \\ \footnotesize{\textbf{($\pm$0.16)}} & & \footnotesize{($\pm$0.21) }\end{tabular}} & \cellcolor{red!9}{\begin{tabular}{ccc}  0.23  & \multirow{2}{*}{\Large{$\backslash$}} & \textbf{0.21} \\ \footnotesize{($\pm$0.16)} & & \footnotesize{\textbf{($\pm$0.16) }}\end{tabular}} \\ \midrule
	\end{tabular}
\caption{Comparison between indicator-based HPO (i.e., IB, columns) and preference-based HPO (i.e., PB, rows). The preference learning model is trained using the non-linear RankSVM implementation and 84 pairwise comparisons.}\label{tbl:end_to_end_evaluation_84_nonlinear}\end{table*}

\begin{table*}[!ht]
\centering
	\begin{tabular}{l|c|c|c|c}
	\toprule
	PB$\backslash$IB & $HV\uparrow$ & $SP\downarrow$ & $MS\uparrow$ & $R2\downarrow$ \\ \midrule
	$HV\uparrow$ & \cellcolor{red!1}{\begin{tabular}{ccc}  0.76  & \multirow{2}{*}{\Large{$\backslash$}} & \textbf{0.77} \\ \footnotesize{($\pm$0.17)} & & \footnotesize{\textbf{($\pm$0.17) }}\end{tabular}} & \cellcolor{blue!15}{\begin{tabular}{ccc} \textbf{ 0.76}  & \multirow{2}{*}{\Large{$\backslash$}} & 0.52 \\ \footnotesize{\textbf{($\pm$0.17)}} & & \footnotesize{($\pm$0.24) }\end{tabular}} & \cellcolor{blue!15}{\begin{tabular}{ccc} \textbf{ 0.76}  & \multirow{2}{*}{\Large{$\backslash$}} & 0.52 \\ \footnotesize{\textbf{($\pm$0.17)}} & & \footnotesize{($\pm$0.21) }\end{tabular}} & \cellcolor{red!1}{\begin{tabular}{ccc}  0.76  & \multirow{2}{*}{\Large{$\backslash$}} & \textbf{0.77} \\ \footnotesize{($\pm$0.17)} & & \footnotesize{\textbf{($\pm$0.16) }}\end{tabular}} \\ \midrule
	$SP\downarrow$ & \cellcolor{blue!67}{\begin{tabular}{ccc} \textcolor{white}{\textbf{ 0.01}}  & \multirow{2}{*}{\textcolor{white}{\Large{$\backslash$}}} & \textcolor{white}{0.03} \\ \footnotesize{\textcolor{white}{\textbf{($\pm$0.01)}}} & & \footnotesize{\textcolor{white}{($\pm$0.02) }}\end{tabular}} & \cellcolor{red!0}{\begin{tabular}{ccc}  0.01  & \multirow{2}{*}{\Large{$\backslash$}} & \textbf{0.01} \\ \footnotesize{($\pm$0.01)} & & \footnotesize{\textbf{($\pm$0.0) }}\end{tabular}} & \cellcolor{blue!100}{\begin{tabular}{ccc} \textcolor{white}{\textbf{ 0.01}}  & \multirow{2}{*}{\textcolor{white}{\Large{$\backslash$}}} & \textcolor{white}{0.04} \\ \footnotesize{\textcolor{white}{\textbf{($\pm$0.01)}}} & & \footnotesize{\textcolor{white}{($\pm$0.03) }}\end{tabular}} & \cellcolor{blue!100}{\begin{tabular}{ccc} \textcolor{white}{\textbf{ 0.01}}  & \multirow{2}{*}{\textcolor{white}{\Large{$\backslash$}}} & \textcolor{white}{0.04} \\ \footnotesize{\textcolor{white}{\textbf{($\pm$0.01)}}} & & \footnotesize{\textcolor{white}{($\pm$0.02) }}\end{tabular}} \\ \midrule
	$MS\uparrow$ & \cellcolor{blue!75}{\begin{tabular}{ccc} \textcolor{white}{\textbf{ 0.62}}  & \multirow{2}{*}{\textcolor{white}{\Large{$\backslash$}}} & \textcolor{white}{0.19} \\ \footnotesize{\textcolor{white}{\textbf{($\pm$0.09)}}} & & \footnotesize{\textcolor{white}{($\pm$0.08) }}\end{tabular}} & \cellcolor{blue!75}{\begin{tabular}{ccc} \textcolor{white}{\textbf{ 0.62}}  & \multirow{2}{*}{\textcolor{white}{\Large{$\backslash$}}} & \textcolor{white}{0.19} \\ \footnotesize{\textcolor{white}{\textbf{($\pm$0.09)}}} & & \footnotesize{\textcolor{white}{($\pm$0.12) }}\end{tabular}} & \cellcolor{red!5}{\begin{tabular}{ccc}  0.62  & \multirow{2}{*}{\Large{$\backslash$}} & \textbf{0.65} \\ \footnotesize{($\pm$0.09)} & & \footnotesize{\textbf{($\pm$0.06) }}\end{tabular}} & \cellcolor{blue!57}{\begin{tabular}{ccc} \textbf{ 0.62}  & \multirow{2}{*}{\Large{$\backslash$}} & 0.23 \\ \footnotesize{\textbf{($\pm$0.09)}} & & \footnotesize{($\pm$0.11) }\end{tabular}} \\ \midrule
	$R2\downarrow$ & \cellcolor{red!4}{\begin{tabular}{ccc}  0.23  & \multirow{2}{*}{\Large{$\backslash$}} & \textbf{0.22} \\ \footnotesize{($\pm$0.17)} & & \footnotesize{\textbf{($\pm$0.16) }}\end{tabular}} & \cellcolor{blue!35}{\begin{tabular}{ccc} \textbf{ 0.23}  & \multirow{2}{*}{\Large{$\backslash$}} & 0.47 \\ \footnotesize{\textbf{($\pm$0.17)}} & & \footnotesize{($\pm$0.23) }\end{tabular}} & \cellcolor{blue!32}{\begin{tabular}{ccc} \textbf{ 0.23}  & \multirow{2}{*}{\Large{$\backslash$}} & 0.45 \\ \footnotesize{\textbf{($\pm$0.17)}} & & \footnotesize{($\pm$0.21) }\end{tabular}} & \cellcolor{red!9}{\begin{tabular}{ccc}  0.23  & \multirow{2}{*}{\Large{$\backslash$}} & \textbf{0.21} \\ \footnotesize{($\pm$0.17)} & & \footnotesize{\textbf{($\pm$0.16) }}\end{tabular}} \\ \midrule
	\end{tabular}
\caption{Comparison between indicator-based HPO (i.e., IB, columns) and preference-based HPO (i.e., PB, rows). The preference learning model is trained using the non-linear RankSVM implementation and 112 pairwise comparisons.}\label{tbl:end_to_end_evaluation_112_nonlinear}\end{table*}

\begin{table*}[!ht]
\centering
	\begin{tabular}{l|c|c|c|c}
	\toprule
	PB$\backslash$IB & $HV\uparrow$ & $SP\downarrow$ & $MS\uparrow$ & $R2\downarrow$ \\ \midrule
	$HV\uparrow$ & \cellcolor{red!3}{\begin{tabular}{ccc}  0.75  & \multirow{2}{*}{\Large{$\backslash$}} & \textbf{0.77} \\ \footnotesize{($\pm$0.18)} & & \footnotesize{\textbf{($\pm$0.17) }}\end{tabular}} & \cellcolor{blue!15}{\begin{tabular}{ccc} \textbf{ 0.75}  & \multirow{2}{*}{\Large{$\backslash$}} & 0.52 \\ \footnotesize{\textbf{($\pm$0.18)}} & & \footnotesize{($\pm$0.24) }\end{tabular}} & \cellcolor{blue!15}{\begin{tabular}{ccc} \textbf{ 0.75}  & \multirow{2}{*}{\Large{$\backslash$}} & 0.52 \\ \footnotesize{\textbf{($\pm$0.18)}} & & \footnotesize{($\pm$0.21) }\end{tabular}} & \cellcolor{red!3}{\begin{tabular}{ccc}  0.75  & \multirow{2}{*}{\Large{$\backslash$}} & \textbf{0.77} \\ \footnotesize{($\pm$0.18)} & & \footnotesize{\textbf{($\pm$0.16) }}\end{tabular}} \\ \midrule
	$SP\downarrow$ & \cellcolor{blue!67}{\begin{tabular}{ccc} \textcolor{white}{\textbf{ 0.01}}  & \multirow{2}{*}{\textcolor{white}{\Large{$\backslash$}}} & \textcolor{white}{0.03} \\ \footnotesize{\textcolor{white}{\textbf{($\pm$0.01)}}} & & \footnotesize{\textcolor{white}{($\pm$0.02) }}\end{tabular}} & \cellcolor{red!0}{\begin{tabular}{ccc}  0.01  & \multirow{2}{*}{\Large{$\backslash$}} & \textbf{0.01} \\ \footnotesize{($\pm$0.01)} & & \footnotesize{\textbf{($\pm$0.0) }}\end{tabular}} & \cellcolor{blue!100}{\begin{tabular}{ccc} \textcolor{white}{\textbf{ 0.01}}  & \multirow{2}{*}{\textcolor{white}{\Large{$\backslash$}}} & \textcolor{white}{0.04} \\ \footnotesize{\textcolor{white}{\textbf{($\pm$0.01)}}} & & \footnotesize{\textcolor{white}{($\pm$0.03) }}\end{tabular}} & \cellcolor{blue!100}{\begin{tabular}{ccc} \textcolor{white}{\textbf{ 0.01}}  & \multirow{2}{*}{\textcolor{white}{\Large{$\backslash$}}} & \textcolor{white}{0.04} \\ \footnotesize{\textcolor{white}{\textbf{($\pm$0.01)}}} & & \footnotesize{\textcolor{white}{($\pm$0.02) }}\end{tabular}} \\ \midrule
	$MS\uparrow$ & \cellcolor{blue!75}{\begin{tabular}{ccc} \textcolor{white}{\textbf{ 0.62}}  & \multirow{2}{*}{\textcolor{white}{\Large{$\backslash$}}} & \textcolor{white}{0.19} \\ \footnotesize{\textcolor{white}{\textbf{($\pm$0.09)}}} & & \footnotesize{\textcolor{white}{($\pm$0.08) }}\end{tabular}} & \cellcolor{blue!75}{\begin{tabular}{ccc} \textcolor{white}{\textbf{ 0.62}}  & \multirow{2}{*}{\textcolor{white}{\Large{$\backslash$}}} & \textcolor{white}{0.19} \\ \footnotesize{\textcolor{white}{\textbf{($\pm$0.09)}}} & & \footnotesize{\textcolor{white}{($\pm$0.12) }}\end{tabular}} & \cellcolor{red!5}{\begin{tabular}{ccc}  0.62  & \multirow{2}{*}{\Large{$\backslash$}} & \textbf{0.65} \\ \footnotesize{($\pm$0.09)} & & \footnotesize{\textbf{($\pm$0.06) }}\end{tabular}} & \cellcolor{blue!57}{\begin{tabular}{ccc} \textbf{ 0.62}  & \multirow{2}{*}{\Large{$\backslash$}} & 0.23 \\ \footnotesize{\textbf{($\pm$0.09)}} & & \footnotesize{($\pm$0.11) }\end{tabular}} \\ \midrule
	$R2\downarrow$ & \cellcolor{red!8}{\begin{tabular}{ccc}  0.24  & \multirow{2}{*}{\Large{$\backslash$}} & \textbf{0.22} \\ \footnotesize{($\pm$0.17)} & & \footnotesize{\textbf{($\pm$0.16) }}\end{tabular}} & \cellcolor{blue!32}{\begin{tabular}{ccc} \textbf{ 0.24}  & \multirow{2}{*}{\Large{$\backslash$}} & 0.47 \\ \footnotesize{\textbf{($\pm$0.17)}} & & \footnotesize{($\pm$0.23) }\end{tabular}} & \cellcolor{blue!29}{\begin{tabular}{ccc} \textbf{ 0.24}  & \multirow{2}{*}{\Large{$\backslash$}} & 0.45 \\ \footnotesize{\textbf{($\pm$0.17)}} & & \footnotesize{($\pm$0.21) }\end{tabular}} & \cellcolor{red!12}{\begin{tabular}{ccc}  0.24  & \multirow{2}{*}{\Large{$\backslash$}} & \textbf{0.21} \\ \footnotesize{($\pm$0.17)} & & \footnotesize{\textbf{($\pm$0.16) }}\end{tabular}} \\ \midrule
	\end{tabular}
\caption{Comparison between indicator-based HPO (i.e., IB, columns) and preference-based HPO (i.e., PB, rows). The preference learning model is trained using the non-linear RankSVM implementation and 140 pairwise comparisons.}\label{tbl:end_to_end_evaluation_140_nonlinear}\end{table*}

\end{document}